\documentclass{article}

\PassOptionsToPackage{numbers, compress}{natbib}
\usepackage[main, final]{neurips_2026}
\usepackage[utf8]{inputenc} % allow utf-8 input
\usepackage[T1]{fontenc}    % use 8-bit T1 fonts
\usepackage{hyperref}       % hyperlinks
\hypersetup{
    colorlinks=true,
    citecolor=blue
}
\usepackage{url}            % simple URL typesetting
\usepackage{booktabs}       % professional-quality tables
\usepackage{amsfonts}       % blackboard math symbols
\usepackage{nicefrac}       % compact symbols for 1/2, etc.
\usepackage{microtype}      % microtypography
\usepackage[dvipsnames]{xcolor}         % colors
\definecolor{grey}{rgb}{0.5,0.5,0.5}
\usepackage{graphicx}
\usepackage{paralist}
\usepackage{wrapfig}

\usepackage{amsmath}
\usepackage{multirow}
\usepackage{gensymb}
\usepackage{enumitem}
\usepackage{makecell}

\usepackage{tcolorbox}
\tcbuselibrary{skins,breakable}
\definecolor{perceivepurple}{HTML}{7952C7}
\definecolor{planteal}{HTML}{008F83}
\definecolor{thengray}{HTML}{737D90}

\DeclareRobustCommand{\PerceiveThenPlan}{%
  \mbox{{\sffamily
    \textcolor{perceivepurple}{\bfseries Perceive}%
    \textcolor{thengray}{\mdseries -then-}%
    \textcolor{planteal}{\bfseries Plan}%
  }}%
}

\title{\PerceiveThenPlan: Layout-as-Policy for Monocular 3D Scene Layout Estimation}
\vspace{-.5em}

\author{
  Junwei Zhou \quad Yu-Wing Tai\thanks{Corresponding author.} \\[0.8em]
  Department of Computer Science \\[0.8em]
  Dartmouth College \\[0.8em]
  \textit{\url{https://colezwhy.github.io/perceivethenplan/}}
}

\begin{document}

\maketitle

\vspace{-.15in}
\begin{figure}[h]
    \centering
    \small
    \includegraphics[width=1\linewidth]{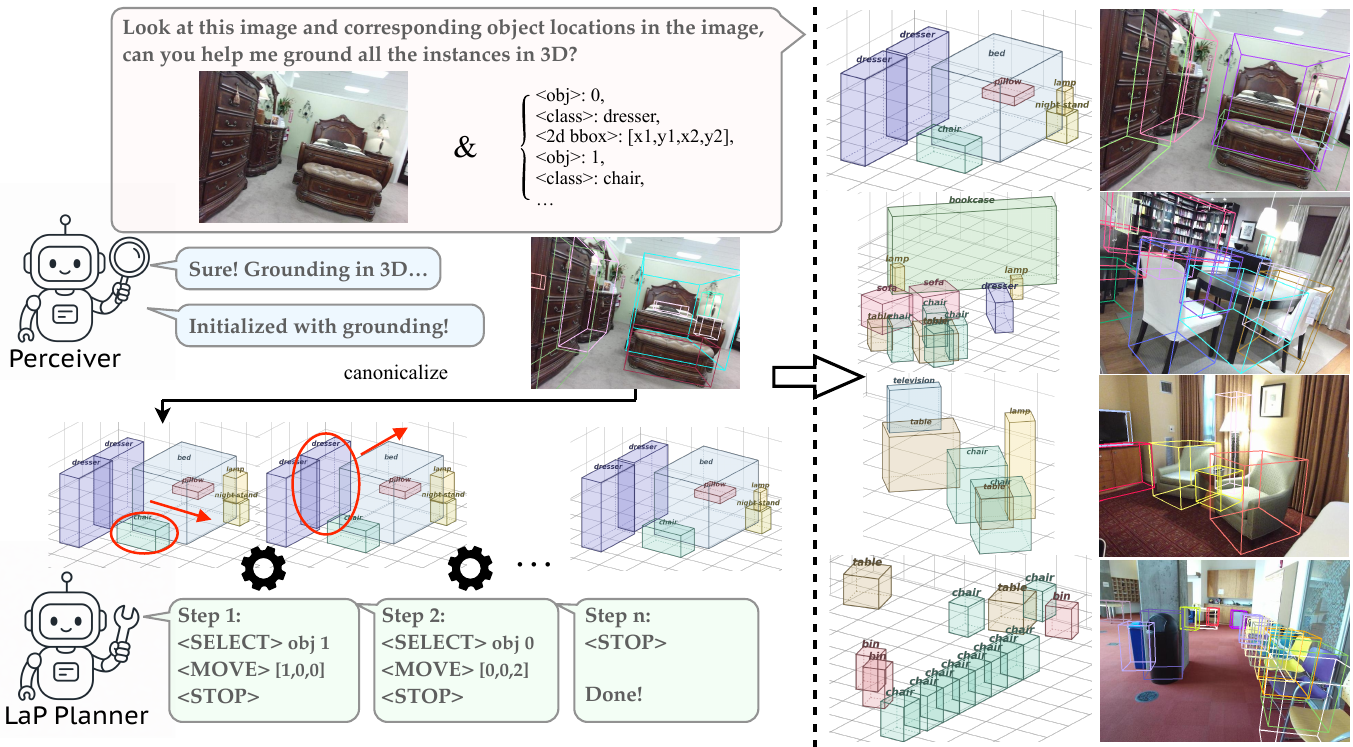}\\
    \vspace{-0.1in}
    \caption{The 3D layout estimation problem can be formulated as a perceive-then-plan problem, where we first ground in 3D with the Perceiver, and then use the LaP Planner to iteratively refine the scene with action sequences following the core design of \textbf{Layout-as-Policy}. } 
    \label{fig:teaser}
    \vspace{.05in}
\end{figure}

\begin{abstract}
% passed
Building structured 3D scene layouts from a single image requires reconciling visual observations with physical and spatial constraints, a challenge that is difficult to address with direct prediction alone. In this work, we formulate monocular 3D layout estimation as a perceive-then-plan problem with vision-language models, where a Perceiver first grounds the 3D objects and then a Planner iteratively refines the scene hypothesis through actions that improve physical plausibility while preserving consistency with the input image. We propose Layout-as-Policy (LaP), which casts the planning stage as a policy learning problem: 3D layouts are represented as structured states, and refined via discrete actions such as translation, rotation, and rescaling. Starting from an observation-aligned initialization with the geometry-enhanced \textbf{Perceiver}, the \textbf{LaP Planner} is trained to produce action sequences that progressively resolve geometric inconsistencies and enforce realistic spatial relations. To enable effective learning, we combine supervised sequence initialization with preference-based optimization, allowing the model to learn corrective behaviors without requiring explicit reward engineering. This formulation transforms layout estimation from a one-shot prediction task into an iterative refinement process, enabling better handling of global constraints and complex object interactions. Experiments demonstrate that our approach produces layouts that are more physically coherent and better aligned with visual observations, while naturally supporting downstream tasks such as scene editing and manipulation.

\end{abstract}

%  main paper input interface 
\section{Introduction}

Recovering structured 3D indoor scene layouts from a single image is a fundamental yet challenging problem, with applications in robotics, simulation, and content creation. Given inherently ambiguous monocular observations, a model must infer object configurations that are not only consistent with the input image but also \textbf{physically plausible} and \textbf{globally consistent}. This requires resolving depth ambiguities, enforcing real-world feasibility (e.g., valid support and non-collision),  and maintaining global coherence across indoor environments.
Most existing approaches formulate monocular 3D layout estimation as a direct prediction problem, where models regress object locations, sizes, and orientations in a single forward pass. While effective in capturing local visual cues, such approaches often struggle to enforce global constraints and inter-object relationships. The underlying challenge is that 3D indoor scenes are inherently structured: object configurations are interdependent, and small local errors can propagate into large global inconsistencies. As a result, one-shot prediction is often insufficient to ensure physically valid and semantically coherent layouts.

In this work, we propose to rethink 3D layout estimation as a perceive-then-plan process with vision-language models (VLMs), where a scene is first initialized by perceiving the input image, and then progressively refined through planning a series of corrective actions. We term this planning stage formulation \textbf{Layout-as-Policy (LaP)}, in which a 3D layout is treated as a scene state, and refinement operations, such as translation, rotation, and rescaling, are modeled as actions governed by a learned policy. With the LaP Planner, we iteratively update the initial layout estimate, gradually resolving inconsistencies and enforcing global constraints while maintaining alignment with the input image. This perspective enables explicit reasoning about constraint satisfaction and provides a natural mechanism for handling realistic object interactions commonly observed in indoor environments.

To realize the Layout-as-Policy framework, we adopt a preference-based policy optimization, that learns to improve layouts through sequential action generation. Rather than relying on explicit reward design, we combine supervised trajectory initialization with direct preference optimization (DPO), enabling the \textbf{LaP Planner} to learn corrective behaviors from relative comparisons between action sequences. This formulation allows the model to capture implicit notions of geometry validity and spatial coherence without requiring handcrafted objective functions.

Another challenge in this paradigm is obtaining a reliable starting point for refinement. To address this, the \textbf{Perceiver} leverages the rich visual and semantic priors of pretrained VLMs to produce observation-aligned initial layouts directly from the image, further enhanced with geometric feature modulation to improve 3D grounding accuracy.
%
% By decoupling perception from planning, our approach combines the strengths of pretrained multimodal models with the flexibility of policy-based optimization, forming a unified system that is both data-efficient and robust.

Together, the Perceiver and the LaP Planner form a closed-loop perceive-then-plan system: the Perceiver provides a visually grounded starting point, and the LaP Planner corrects its physical and relational errors through policy-driven action planning. This 
decomposition makes each stage focus on what it does best, with the 
Planner's action-based interface naturally enabling applications such as language-guided scene editing and embodied manipulation.

We summarize our contributions as follows:
\begin{compactitem}
\item We introduce a \emph{perceive-then-plan} pipeline for monocular 3D layout estimation with vision-language models, and propose \textbf{Layout-as-Policy (LaP)} as our core formulation, which casts the planning stage as policy learning over structured scene states and discrete actions with the preference-optimized \textbf{LaP Planner}.
\item We design a geometry-enhanced \textbf{Perceiver} that integrates visual and geometric features to produce reliable observation-aligned initial layouts, providing a strong starting point for action-based refinement.
\item Experiments demonstrate that iterative action-based refinement yields layouts with improved structural coherence and visual alignment, while the action-based interface of \textbf{LaP} naturally supports downstream tasks such as scene editing and embodied planning, opening avenues for future extensions.
\end{compactitem}

\section{Related Works}
\label{sec:related}
% Too long right now. Shorterning.
\subsection{VLMs for Spatial Reasoning and Scene Representation}
Recent advances in foundational Vision-Language Models (VLMs)~\cite{liu2023llava, hurst2024gpt, tong2024cambrian1fullyopenvisioncentric, liu2025agentic3dscenegeneration} have enabled strong performance in visual grounding~\cite{li2024llava, bai2025qwen25vl, weaksam, dai2023instructblip} and compositional spatial reasoning~\cite{hong20233dllm, kim2024openvla, gholami2025spatialreasoningvisionlanguagemodels,  cai2026idealbenchindoordatasetevaluation, liu2024chatcam}, with a growing body of work using structured language as an intermediate representation for 3D scenes~\cite{zha2025enable, yang2025thinking, zhu2024llava3d, wang2025n3d, chen2024spatialvlm, li2025seeground, xu2024vlmgrounder}.
More recently, VLMs are applied to structured 3D scene modeling and generation~\cite{gu2025artiscene, liu2025worldcraft, ling2025scenethesislanguagevisionagentic, zhang2025scenelanguagerepresentingscenes, motionagent}, formalizing 3D environments as executable programs to enable higher-level reasoning and manipulation.
Despite these advances, existing approaches largely decouple visual grounding from spatial reasoning and planning, and rely on purely language-centric VLMs without leveraging explicit geometric cues. In contrast, our \textbf{Layout-as-Policy (LaP)} formulation unifies the two under a \emph{perceive-then-plan} paradigm: a geometry-enhanced {Perceiver} grounds 3D by fusing visual and geometric features, and a {LaP Planner} iteratively refines the layout through policy-driven actions, coupling perception and reasoning for globally coherent 3D layouts.

\subsection{Monocular 3D Scene Layout Estimation and Synthesis}
Monocular 3D scene understanding has long been studied through object detection, layout estimation, and holistic reconstruction, leveraging geometric priors, contextual cues, and learned depth~\cite{wang2015holistic, zhou2019holistic, nie2020total3dunderstanding, wang20193droomnet, zhang2021holisticimplicit, zhang2021deepPanocontext}.
In parallel, 3D scene synthesis generates structured environments from visual or textual inputs, spanning scene reconstruction~\cite{zhou2026gena, chen2025sam, wu2025amodal3r, meng2026scenegen}, composed generation~\cite{zhou2025coco, zhou2024gala3d, zhou2024layout, wang2025tabletopgen}, and controllable, compositional, or physics-aware synthesis~\cite{chen2023control3d, ge2025compgs, chen2024comboverse, cao2025physx}. More recently, VLMs have been adopted for 3D layout prediction~\cite{li2024advances, gu2025artiscene, Yang_2024_CVPR}, operating over 2D bird's-eye-view layouts~\cite{sun2024layoutvlm, ran2025direct, ccelen2024design} or full 3D parameterizations~\cite{feng2023layoutgpt, liu2025worldcraft}.
Both lines predominantly adopt \emph{one-shot, feedforward} formulations, making it hard to correct intermediate errors or enforce non-collision and support constraints. In contrast, our \textbf{Layout-as-Policy} (LaP) framework casts monocular layout estimation as an iterative \emph{perceive-then-plan} process: a Perceiver grounds 3D objects, and a LaP Planner refines the layout via policy-driven actions over structured scene states.

\subsection{Sequential Decision-Making and Policy Learning}
Sequential decision-making has emerged as a powerful paradigm where outputs are interdependent and global consistency is required. Classical reinforcement learning formulates such problems as policies over structured states~\cite{sutton2018reinforcement}, a perspective now extended to multimodal reasoning through post-training alignment of VLMs~\cite{tie2025survey, cao2024survey}: supervised fine-tuning (SFT) adapts models via instruction-response data~\cite{zhang2025instructiontuninglargelanguage, li2024llava}, while RLHF~\cite{bai2022training, guo2025deepseek} and preference-based optimization~\cite{rafailov2023direct, ethayarajh2024kto} align behaviors through relative comparisons without explicit reward modeling. Building on these advances, action-based reasoning and agentic frameworks further treat LLMs/VLMs as policies over structured environments~\cite{yao2022react, xu2025llava, hu2024scenecraftllmagentsynthesizing}.
Yet these approaches target open-ended reasoning or embodied interaction, not structured 3D layout estimation. Our \textbf{Layout-as-Policy} (LaP) formulation fills this gap: a layout is a structured state, refinement operations are actions, and a LaP Planner trained via SFT followed by DPO iteratively improves layouts while maintaining visual consistency.
\section{Method}
\label{sec:method}
\begin{figure}[t]
    \centering
    \small
    \includegraphics[width=1\linewidth]{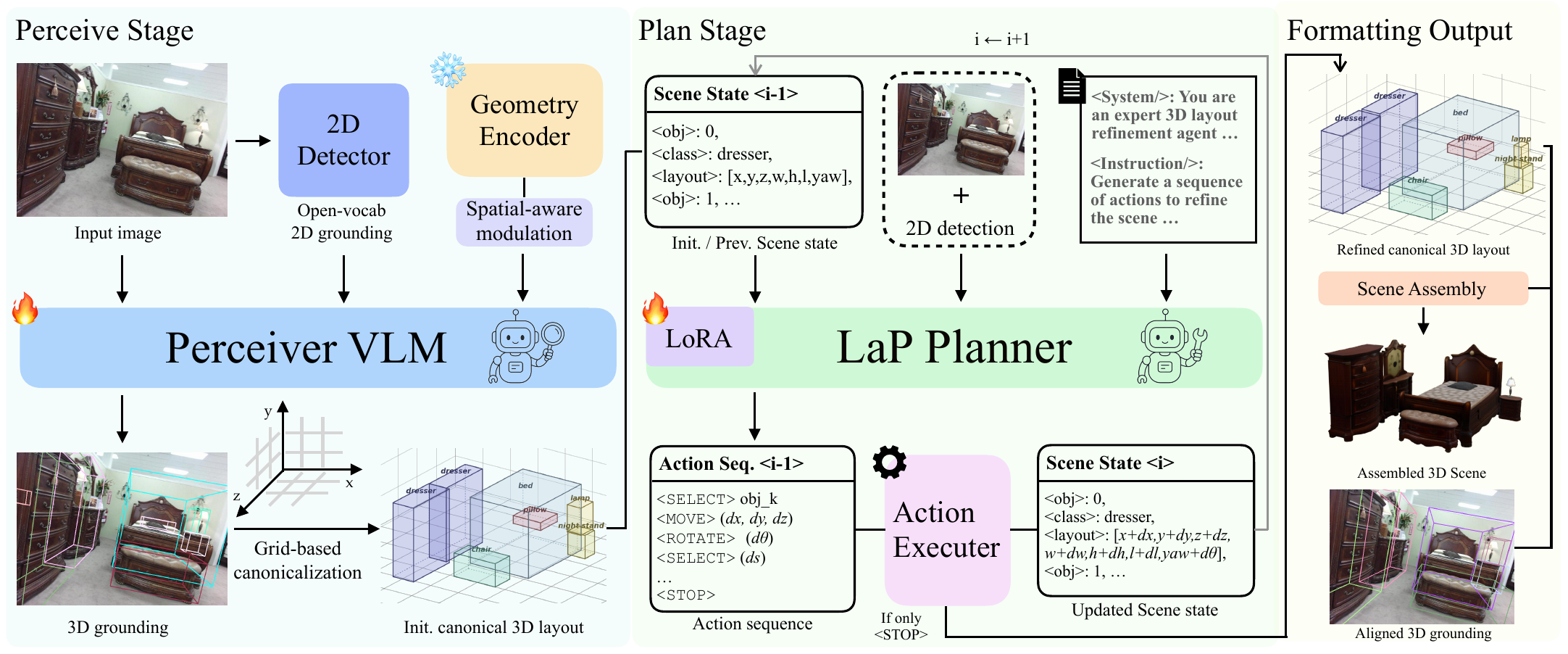}\\
    \vspace{-0.05in}
    \caption{Overview of our \emph{perceive-then-plan} framework. The Perceiver grounds 3D boxes from the input image. A canonicalized, grid-based representation is then iteratively refined by the LaP Planner, which treats each layout as a structured state and selects discrete actions (translate, rotate, rescale) via a learned policy until convergence. The final layout supports both scene assembly (digital twin) and camera-space projection (3D grounding).} 
    \vspace{-0.1in}
    \label{fig:pipeline}
\end{figure}

Given a single image $I$ and a set of 2D detections $\mathcal{B}^{2d} = \{b_i^{2d}\}_{i=1}^{N}$, our goal is to recover a structured 3D scene layout $\mathcal{L} = \{l_i\}_{i=1}^{M}$ that describes the spatial arrangement of objects in the scene. Each 2D detection $b_i^{2d} = (c_i, x_i, y_i, w_i, h_i)$ consists of a semantic category $c_i$ and a 2D bounding box parameterized by its center coordinates and dimensions. Each element $l_i$ in the output layout represents a 3D object, including its semantic category, 3D position, 3D dimensions, and orientation.

We propose a perceive-then-plan two-stage pipeline that decomposes this task into \textbf{3D grounding} and \textbf{layout refinement}. In the Perceive Stage, a Perceiver VLM takes the image $I$ and 2D detections $Box_{2d}$ as input to produce an initial 3D groundings  $Box_{3d}$ by lifting each detected object into 3D space, the $Box_{3d}$ is then canonicalized with our grid-based representation to achieve canonical 3D layout $\hat{\mathcal{L}}_{3d}$. In Plan Stage, a LaP Planner takes $I$, ${Box}_{2d}$, and $\hat{\mathcal{L}}_{3d}$ as input and iteratively refines it through a sequence of structured actions (e.g., moving, rotating) to manipulate objects to produce the final structurally coherent layout $\mathcal{L}^*_{3d}$.

\subsection{Grounding as Initialization}
\label{sec: 31vlmground}
\noindent\textbf{Preliminary.} Given the input image $I$ and corresponding 2D detection results $Box_{2d}$, our goal is to ground 3D boxes $Box_{3d}$ of the 2D detection results in the image. We use a vision-language model (denoted as \textbf{Perceiver}) for this task because of the strong open-world knowledge, image understanding, and structured output ability (See Appendix~\ref{3dgrounding} for design choice). 
Intuitively, a single input image is the easiest way to convey the model a 3D world, and the major difficulty for a model to understand the 3D scene hidden behind is to disregard all the camera pose shifts since not all the images are taken in the canonical pose.
To overcome this difficulty, the common way is to directly regress the vertices of 3D boxes. However, this regression requires strong format priors, which does not fit the optimization target of vision-language models. Thus we use a \emph{local-axis coordinate representation} instead to address the learning gap. For each 3D box, we mark the two bottom local axes as $ax_{x}$ and $ax_{z}$, and the vertical axis $ax_{y}$ is given by a standard cross product, respectively:
\begin{align}
    \label{eq1:crossup}
    ax_{y} &= ax_{x} \times ax_{z}. 
\end{align}
Together, for each 3D box, we can obtain its parameterized representation: 
\begin{align}
    \label{eq2:box3d}
    Box_{3d} = \{(x_i, y_i, z_i, w_i, h_i, l_i, ax_{x}, ax_{z}, cls_i)\}_{i=1}^N,
\end{align}
where $x_i, y_i, z_i$ are the center coordinates, $w_i, h_i, l_i$ are the scales, $N$ is the number of objects, and $cls_i$ is the class label, respectively.
This representation serves as the grounding target in stage 1.

\noindent\textbf{Geometry-aware Feature Modulation.}
Directly fine-tuning the VLM aligns its output format but only marginally improves 3D grounding. We therefore introduce a lightweight geometry-aware modulation module that injects geometric priors into the VLM's visual features.
Given 2D features $F_{2d}\in\mathbb{R}^{H\times W\times C}$ from the VLM vision tower and geometry features $F_{3d}$ from a pretrained encoder (i.e., VGGT~\cite{wang2025vggt}), we regress FiLM~\cite{perez2018film} coefficients and a spatial gate from $\tilde{F}_{3d}=\mathrm{LN}(F_{3d})$:
\begin{align}
    \label{eq:geometrymodulation}
    F_{\mathrm{fused}} = g\odot\big[(1+\Delta\gamma)\odot F_{2d} + \beta\big] + (1-g)\odot F_{2d},
\end{align}
where $\Delta\gamma,\beta=\mathrm{MLP}_{\gamma\beta}(\tilde{F}_{3d})$, $g=\sigma(\mathrm{MLP}_{g}(\tilde{F}_{3d}))\in\mathbb{R}^{H\times W\times 1}$, and $\odot$ denotes element-wise multiplication. The residual form $(1+\Delta\gamma)$ keeps $F_{\mathrm{fused}}\!\approx\!F_{2d}$ at initialization, while $g$ adaptively controls per-location modulation strength.

\noindent\textbf{Grid-based Representation in Language Space.}
To canonicalize the grounded 3D boxes into a layout amenable to the subsequent action-based refinement, we discretize them into a grid-based representation in a gravity-aligned frame.
We first estimate the gravity direction $\mathbf{g}$ as the dominant axis from an SVD of all bottom-face normals, and construct an orthonormal frame
$\mathbf{y}=\mathbf{g}$, $\mathbf{x}=\operatorname{proj}_{\perp\mathbf{y}}(\mathbf{e}_x)$, $\mathbf{z}=\mathbf{x}\times\mathbf{y}$,
where $\mathbf{e}_x$ is the normed camera right axis. All vertices are rotated by $\mathbf{R}_{\mathrm{ga}}=[\mathbf{x};\mathbf{y};\mathbf{z}]^\texttt{T}$ and translated so the scene is horizontally centered with the ground plane, yielding parameterized and standard layout boxes $(x,y,z,w,h,l,\theta)$. Benefiting from the structural priors of VLMs, the generated layouts are already consistent with the ground plane, resulting in negligible conversion error ($\sim 10^{-4}$).
Each continuous value is then discretized into integer grid indices:
\begin{align}
  \label{eq:grid}
  g = \operatorname{round}\!\left(\tfrac{v - o}{\delta}\right), \quad
  g_\theta = \operatorname{round}\!\left(\tfrac{\theta + \pi}{2\pi / N_\theta}\right),
\end{align}
where $\delta$ (e.g., $0.1$m) is the cell size, $o$ is a per-scene offset ensuring non-negative indices, and $N_\theta$ is the number of yaw bins. 
Finally, we can achieve initial canonicalized 3D layout $\hat{\mathcal{L}}_{3d}$.

\subsection{Action-based Layout Refinement}
% \vspace{-.05in}
\label{sec: 32actionrefinement}
After the canonical 3D layout $\hat{\mathcal{L}}_{3d}$ is initialized, we treat the layout estimation problem as policy learning as our core insight, denoted as \textbf{Layout-as-Policy (LaP)}. We adopt a LaP Planner VLM trained with LoRA~\cite{hu2022lora} to iteratively refine the 3D layout $\hat{L}$ toward a layout state $\mathcal{L}^*$ that is both physically coherent and visually aligned with the action sequences that are iteratively applied to update the scene state.
In this section, we first turn the initial layout $\hat{\mathcal{L}}_{3d}$ into language-based scene states, where each entity (layout box) can be refined with action sequences generated by the VLM. We iteratively refine the scene state and apply action sequences with a feedback loop until it reaches a stable state, which results in a visually aligned and physically plausible 3D layout set.

\begin{wrapfigure}{r}{0.4\textwidth}
\vspace{-.1in}
  \begin{centering}
\includegraphics[width=0.4\textwidth]{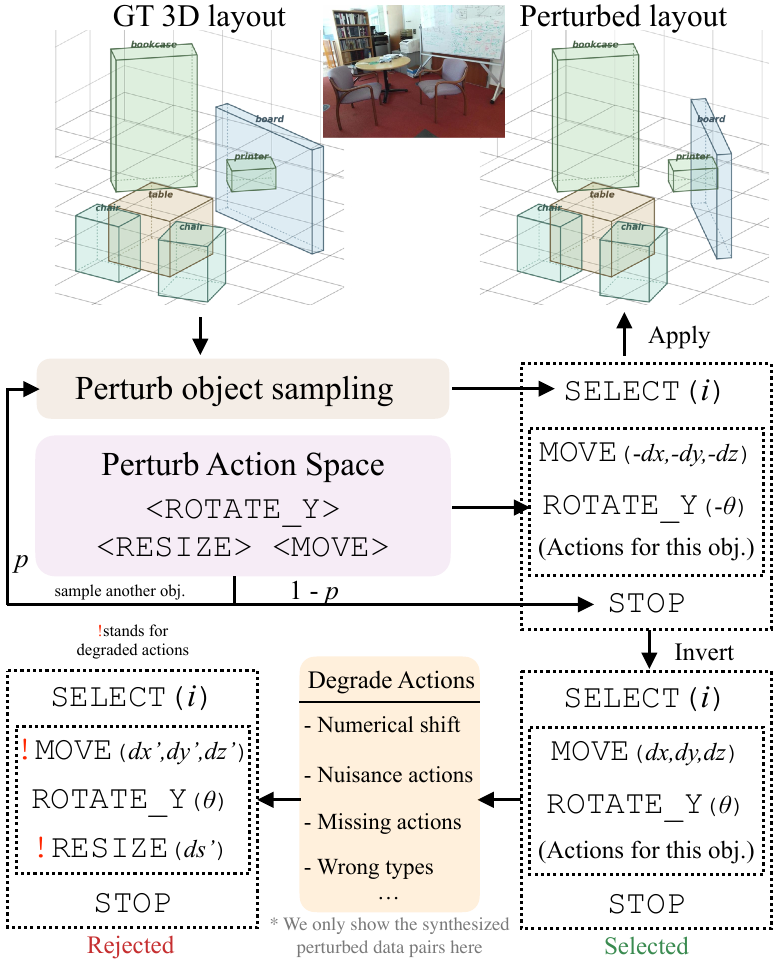}
  \end{centering}\\
  \vspace{-.2in}
  \caption{\footnotesize A simple illustration of DPO data pair synthesis. We note that the action sequences can support multiple objects.}
  \label{fig:dpodata}
  \vspace{-.7in}
\end{wrapfigure}

\noindent\textbf{Action Space.} We design a compact action space that captures the minimal yet sufficient set of operations for layout refinement. Each action is formulated in a programming language format with explicit parameters:
% \vspace{-.5em}
\begin{description}[style=unboxed, leftmargin=2em]
    \item[\texttt{SELECT}($i$)] Selecting the object $i$ in the structured layout $\mathcal{L}_{3d} = \{l_i\}_{i=1}^{M}$ for later actions.
    \item[\texttt{MOVE}($dx, dy, dz$)] Move the object along the ground plane by $dx$ and $dz$ and vertically by $dy$.
    \item[\texttt{RESIZE}($ds$)] Rescale all three dimensions of the bounding box by a factor of $(1 + ds*0.1)$.
    \item[\texttt{ROTATE\_Y}($d\theta$)] Rotate the object around the Y axis by $\theta$ degrees, adjusting its yaw orientation.
    \item[\texttt{STOP}] Stop the current action sequence for output sequence.
\end{description}
% \vspace{-.5em}
These primitives target the dominant failure modes of the Perceiver, i.e., positional misalignment, incorrect vertical placement, inaccurate scale, and erroneous orientation. Iteratively composing them allows the refinement model to resolve inter-object collisions and enforce ground contact, improving the physical and structural plausibility of the predicted layout.

% here originally

\noindent\textbf{Perturbed Data Synthesis.} 
We synthesize perturbed 3D layouts to simulate grounding initializations that are generally correct but contain minor physical inconsistencies and image misalignments. 
As shown in Fig.~\ref{fig:dpodata}, starting from a scene's ground-truth layout, we sample an object and randomly draw non-repeating perturbations from the action space; with probability $p$ we continue sampling another object, and with probability $1-p$ we 
stop. 
The accumulated perturbations are applied to produce the perturbed layout used as training input, and we invert these perturbation sequences to obtain the ground-truth action sequence. The synthesis can simulate a wide range of conflicts of the initialized scene (See Appendix~\ref{app:actionseq}). 

\noindent\textbf{Supervised Fine-Tuning Warm-up.}
To equip the LaP Planner with the instruction-aligned output format and a basic action-layout-image comprehension ability, we first perform supervised fine-tuning (SFT) with the synthesized perturbed data. We take the perturbed 3D layout and the ground truth action sequence for constructing training format.
% %
% We construct training samples by pairing perturbed layouts with their corresponding ground-truth action trajectories in record, which can recover the original layout through back-applying actions to the perturbed layout. 
%
During training, we use a system prompt for background information and a user prompt instructing the model to generate corrective action sequences, conditioned on the input image. 
In this warm-up stage, the Planner learns to produce reasonable refinement trajectories that reduce spatial artifacts and improve alignment with visual observations and geometric consistency in a progressive way, instead of directly learning from noisy Perceiver-initialized data.

\noindent\textbf{Refinement as Preference Learning.}
After the SFT warm-up, we observe that the action-based decision-making framework is particularly well-suited for preference learning: 
the refinement task is inherently comparative, where judging which action leads to a better layout is far easier than defining a single optimal one, and the discrete action space enables natural pairwise construction.
Intuitively, we apply Direct Preference Optimization (DPO)~\cite{rafailov2023direct} using offline-synthesized trajectory pairs (Fig.~\ref{fig:dpodata}). 
The preference data consists of trajectories from two sources: Perceiver-initialized action sequences and manually perturbed variants (See Appendix~\ref{app:laptraining}). 
We treat the ground-truth action sequences toward the canonicalized layouts as $\texttt{selected}$ samples, while both Perceiver-initialized and perturbed trajectories are used as $\texttt{rejected}$ samples, covering a wide range of inconsistencies.
After that, we use the standard DPO training framework on the LaP Planner, teaching the model not only to learn how to apply actions, but also to implicitly understand actions from diverse pair-wise comparisons.

\noindent\textbf{Iterative Refinement.}
Correcting all layout inconsistencies at one time is challenging due to the combinatorial decision space. We therefore refine the layout iteratively over multiple rounds. Let $\mathcal{L}_{3d}^0 = \{l_i\}_{i=1}^{M}$ be the initial layout from the Perceiver. At round $t$, the Planner takes $\mathcal{L}_t$ and the input image as context, predicts an action sequence $\mathcal{A}_t$, and applies it to obtain $\mathcal{L}_{3d}^{t+1} = \texttt{Apply}(\mathcal{L}_{3d}^t, \mathcal{A}_{3d}^t)$. Repeating for $n$ rounds yields a trajectory $\mathcal{L}_{3d}^0, \mathcal{L}_{3d}^1, \ldots, \mathcal{L}_{3d}^n$, where each layout is expected to be more physically plausible than its predecessor, and we denote the final layout as ${\mathcal{L}}_{3d}^*$. This decomposition lets the model resolve a subset of violations per round rather than all at once, leading to more stable convergence ($\texttt{<STOP>}$ only or reach the round limit).

\vspace{-1em}
\subsection{Scene Assembly}
\vspace{-.5em}
\label{sec: assembly}
The refinement of plan stage outputs a set of 3D layout boxes, but real 3D assets retrieved or generated from these boxes rarely match them exactly, since shape variations and misaligned contact surfaces can introduce floating or penetration artifacts upon placement.
We therefore apply a minor post-processing step to enable realistic scene assembly. With the LoRA adapter disabled, the LaP Planner generates a contact scene graph from the input image, where each edge encodes a pairwise support relation. 
Objects and their supporters are grouped into relational bundles, and a single-pass gravity simulation is performed on these bundles under floor constraints. Operating on bundles corrects placement artifacts while preserving the relational structure established during refinement.
% \vspace{-.1in}
\section{Experiments}
\label{sec:exp}
% \vspace{-.05in}
\subsection{Experimental Settings}
% \vspace{-.05in}
\noindent\textbf{Dataset.} 
We use the training and validation splits of two indoor 3D layout datasets, Hypersim~\cite{roberts2021hypersim} and SUN RGB-D~\cite{song2015sun}, for training. 
For Perceiver training, we project ground-truth 3D bounding boxes onto the image plane and retain only their visible regions as input 2D boxes to simulate amodal grounding. 
We further filter scenes containing fewer than 5 or more than 30 objects, resulting in 42{,}160 training samples. 
For the LaP Planner, we construct supervised fine-tuning (SFT) and DPO training data using the grid-based canonicalization and controlled perturbation strategy illustrated in Fig.~\ref{fig:dpodata}. 
Finally, we randomly sample 1,000 image-scene pairs from the SUN RGB-D test split for evaluation. Additional implementation details are provided in Appendix~\ref{app:imple}.

% \vspace{-.2em}
\noindent\textbf{Perceiver.} We compare our Perceiver against state-of-the-art VLMs, both 
open-source (LLaMA3.2-Vision~\cite{grattafiori2024llama3herdmodels}, Qwen-3-VL~\cite{bai2025qwen3}, GLM-4.6V-Flash~\cite{5team2025glm45agenticreasoningcoding}, InternVL3.5~\cite{chen2024internvl}, Gemma3~\cite{gemmateam2025gemma3technicalreport}, N3D-VLM~\cite{wang2025n3d}) and closed-source (Gemini~2.5-Flash~\cite{geminiteam2025geminifamilyhighlycapable}, GPT-4o~\cite{hurst2024gpt}), under two settings: \emph{image + 2D detections} and \emph{direct 3D grounding}. 
We report Reprojection IoU, Precision@IoU=25/50, and averaged 
Depth Error to evaluate the grounding precision (See Appendix~\ref{app:eval} for details). 

\noindent\textbf{LaP Planner.} Since currently there aren't any baselines working on VLMs for 3D layout/grounding refinement to compare with, we evaluate the LaP Planner by comparing layouts \emph{before} and \emph{after} refinement, which isolates the contribution of action-based iterative correction from upstream perception errors. To validate the effectiveness of our action-based paradigm, we also compare with \textit{trained one-shot VLM-based direct layout refinement}, \textit{rule-based refinement}, and GPT-4o iterative.
We evaluate along: 1) visual alignment \& geometry consistency, measured by Reproj. IoU and Avg. Depth Error, to verify that refinement does not drift away from image evidence; 2) physical plausibility, measured by collision count (pairwise 3D box penetration), Support Violation Rate (SVR, floating or penetrated objects lacking valid support), and averaged Rotation Error of all objects in the scene, which directly quantify the geometric constraints our Planner is trained to enforce. Full metric definitions and implementations of methods compared with our LaP Planner are in Appendix~\ref{app:complap}.

{\begin{table}[t!]
\centering
\caption{We evaluate the 3D grounding ability of our Perceiver. We compare our method with both open-source and closed-source models. The best and second best results are \textbf{bold} and \underline{underlined}. \ddag{} stands for using simple supervised fine-tuning and \dag{} represents direct 3D grounding.}
\resizebox{\textwidth}{!}
{
\begin{tabular}{l|l|cccc}
\toprule
Category & Model & Reproj. IoU$\uparrow$ & Prec.{\tiny@(IoU=0.25)} $\uparrow$ & Prec.{\tiny@(IoU=0.5)} $\uparrow$ & Avg. DE$\downarrow$ \\
\midrule
\multirow{2}{*}{Closed-source} & Gemini-2.5-flash\dag & 0.24 & 0.19 & 0.07 & 0.47  \\ 
                               & GPT-4o\dag & 0.27 & 0.19 & 0.08  & 0.55  \\ 
\midrule
\multirow{9}{*}{Open-source}    & LLaMA3.2-Vision 11B & 0.14 & 0.18 & 0.09 & 0.54 \\
                                & Qwen3-VL 8B & 0.18 & 0.31 & 0.13  & 0.43  \\ 
                                & Qwen3-VL 8B\ddag & 0.44 & 0.84 & 0.46  & 0.18  \\ 
                                & Qwen3-VL 8B\dag & 0.31 & 0.54 & 0.28  & 0.25  \\ 
                                & Qwen3-VL 32B & 0.26 & 0.32 & 0.18 & 0.37  \\ 
                                & Qwen3-VL 32B\dag & 0.33 & 0.57 & 0.29 & 0.21  \\ 
                                & InternVL3.5 8B & 0.13 & 0.24 & 0.11 & 0.58  \\ 
                                & Gemma3 27B & 0.24 & 0.23 & 0.14 & 0.47  \\ 
                                & Gemma3 27B\dag & 0.30 & 0.22 & 0.15 & 0.55  \\ 
                                & GLM-4.6V-Flash 9B & 0.21 & 0.28 & 0.13 & 0.36  \\
                                & N3D-VLM 7B\dag       & 0.33 & 0.53 & 0.35 & 0.19  \\

\midrule
\multirow{2}{*}{Ours}          & Perceiver 2B & \underline{0.54} & \textbf{0.91} & \underline{0.62} &                                          \underline{0.09} \\     
                                & Perceiver 8B & \textbf{0.58} & \underline{0.90} & \textbf{0.64} & \textbf{0.08}  \\   
\bottomrule
\end{tabular}
 }
\label{tab: stage1grounding}
\vspace{-.05in}
\end{table}
}
%\vspace{-.1in}

% \begin{table}[ht!]
% \centering
% \small
% \caption{The quantitative results on the 3D grounding task in our stage 1 initialization. We compare our method with both open-source and closed-source models. The best and second best results are \textbf{bold} and \underline{underlined}. \ddag~stands for with the same format of supervised fine-tuning.}
% \begin{tabular}{l|l|cccc}
% \toprule
% \multirow{2}{*}{Category} & \multirow{2}{*}{Model} 
%   & \makecell{Reproj\\IoU$\uparrow$} 
%   & \makecell{Precision$\uparrow$\\{\tiny @IoU=0.25}} 
%   & \makecell{Precision$\uparrow$\\{\tiny @IoU=0.5}} 
%   & \makecell{Average\\Depth Error$\downarrow$} \\
% \midrule
% \multirow{2}{*}{Closed-source} & Gemini-3-flash & -- & -- & -- & --  \\ 
%                                & ... & -- & -- & -- & --  \\ 
% \midrule
% \multirow{5}{*}{Open-source}   & Qwen3-VL 8B & 0.22 & 0.20 & -- & 0.61  \\ 
%                                 & Qwen3-VL 8B\ddag & 0.36 & 0.20 & -- & 0.18  \\ 
%                                 & Qwen3-VL 32B & 0.26 & 0.22 & -- & 0.60  \\ 
%                                 & InternVL3.5 & -- & -- & -- & --  \\ 
%                                 & ... & -- & -- & -- & --  \\ 
% \midrule
% \multirow{2}{*}{Ours}          & Perceiver 2B & \underline{0.58} & \underline{0.71} & -- &                                          \textbf{0.05} \\     
%                                 & Perceiver 8B & \textbf{0.61} & \textbf{0.72} & -- & \textbf{0.05}  \\   
% \bottomrule
% \end{tabular}
% \label{tab:stage1grounding}
% \end{table}
\begin{table}[t!]
\centering
\small
\setlength{\tabcolsep}{6pt}
\caption{Numerical results of action-based refinement by the LaP Planner. We use the 2B Perceiver as initialization. {\color{red}Red} numbers shows performance degradation and {\color{ForestGreen}green} shows gain.}
\label{tab:stage2_refinement}
\resizebox{\textwidth}{!}{
\begin{tabular}{l|l ccccc}
\toprule
Model & State & Reproj. IoU $\uparrow$ & SVR (\%) $\downarrow$ & \# Collisions $\downarrow$ & Rot. Err. $\downarrow$ & Avg. DE$\downarrow$ \\
\midrule
Perceiver only      & Before Ref.          & 0.54 & 11.63 & 3.62 & 17.74 & 0.09 \\
\midrule
One-shot VLM direct Ref.   & After Ref.           & 0.55{\scriptsize\color{ForestGreen}+0.01}   & 10.48{\scriptsize\color{ForestGreen}-1.15}&  3.54{\scriptsize\color{ForestGreen}-0.08}& 14.83{\scriptsize\color{ForestGreen}-2.91}& 0.11{\scriptsize\color{red}+0.02}\\
Rule-based Ref.     & After Ref.           & 0.26{\scriptsize\color{red}-0.28}   & 0.0{\scriptsize\color{ForestGreen}-11.63}& 0.0{\scriptsize\color{ForestGreen}-3.62}& 17.74{\scriptsize\color{gray}+0.0}& 0.24{\scriptsize\color{red}+0.15}\\
GPT-4o             & After Ref. (converge)  & 0.49{\scriptsize\color{red}-0.05}   & 9.96{\scriptsize\color{ForestGreen}-1.67}& 3.89{\scriptsize\color{red}+0.27}& 13.35{\scriptsize\color{ForestGreen}-4.39}& 0.13{\scriptsize\color{red}+0.04}\\
\midrule
\multirow{2}{*}{LaP Planner (2B)}
                    & After Ref. (1 iter)  & 0.57{\scriptsize\color{ForestGreen}+0.03} & 8.81{\scriptsize\color{ForestGreen}-2.82}& 3.21{\scriptsize\color{ForestGreen}0.41}& 15.25{\scriptsize\color{ForestGreen}-2.49}& 0.09{\scriptsize\color{gray}+0.0}\\
                    & After Ref. (converge)& {0.65}{\scriptsize\color{ForestGreen}+0.11} & {4.85}{\scriptsize\color{ForestGreen}-6.78}& {1.46}{\scriptsize\color{ForestGreen}-2.16}& {9.05}{\scriptsize\color{ForestGreen}-8.69}& {0.07}{\scriptsize\color{ForestGreen}-0.02}\\
\midrule
\multirow{2}{*}{LaP Planner (8B)}
                    & After Ref. (1 iter)  & 0.57{\scriptsize\color{ForestGreen}+0.03} & 8.62{\scriptsize\color{ForestGreen}-3.02}& 3.23{\scriptsize\color{ForestGreen}-0.39}& 14.37{\scriptsize\color{ForestGreen}-3.37}& 0.09{\scriptsize\color{gray}+0.0}\\
                    & After Ref. (converge)& 0.67{\scriptsize\color{ForestGreen}+0.13} & 4.12{\scriptsize\color{ForestGreen}-7.52}& 1.44{\scriptsize\color{ForestGreen}-2.18}& 8.62{\scriptsize\color{ForestGreen}-9.12}& 0.07{\scriptsize\color{ForestGreen}-0.02}\\
\bottomrule
\end{tabular}
}
\vspace{-.1in}
\end{table}

\begin{figure}[t!]
    \centering
    \small
    \includegraphics[width=1\linewidth]{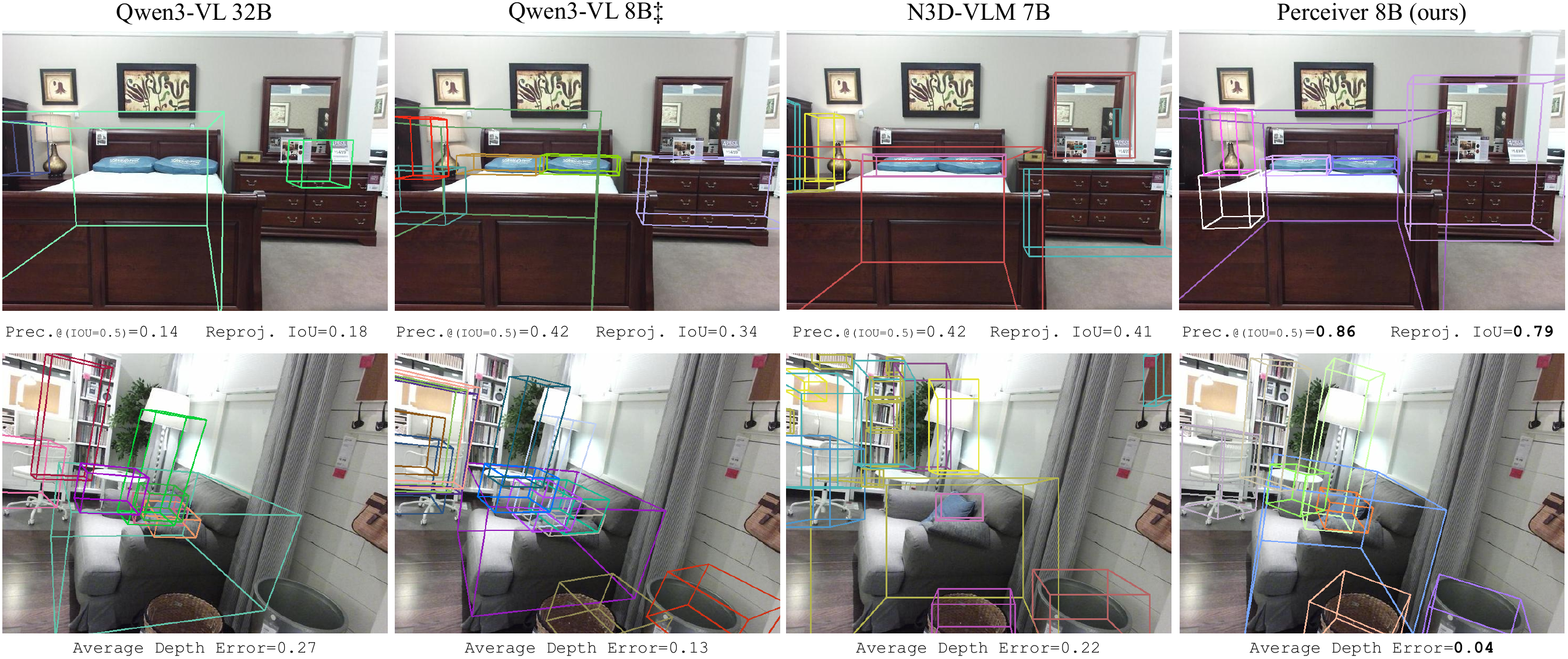}\\
    % \vspace{-0.05in}
    \caption{We visualize the 3D grounding result generated by our Perceiver 8B model. For better demonstration, we set each instance a different color in the visualization and provide the numerical metrics in each sample for direct comparison.} 
    \vspace{-0.05in}
    \label{fig:perceiver}
\end{figure}

\subsection{Main Results}
% We present both quantitative and qualitative comparisons for the 
% Perceiver and the LaP Planner, evaluating each component using the metrics defined above.

\noindent\textbf{Quantitative Results.} As shown in Tab.~\ref{tab: stage1grounding}, our Perceiver outperforms existing VLMs and VLM-based grounding methods by a large margin under 
both evaluation settings, confirming the effectiveness of our 
geometry-enhanced design for monocular 3D lifting. 
Tab.~\ref{tab:stage2_refinement} further compares layouts before 
and after LaP Planner refinement: we observe substantial drops in 
collision count and Support Violation Rate together with improved 
Reprojection IoU, validating our \textbf{Layout-as-Policy} formulation 
and action-based refinement design. It is worth noting that \textit{trained one-shot VLM-based direct refinement}, with the same amount of training data as our LaP Planner, do not achieve satisfying results, and even add to the depth error (We also present using one-shot VLM direct refinement for multiple-rounds in Appendix~\ref{app:complap}). 
While \textit{rule-based refinement} alleviates physical inconsistencies, it suffers from poor visual alignment due to the absence of image-conditioned reasoning. 
These results further highlight the necessity of our iterative action-based refinement framework.

\begin{figure}[t!]
    \centering
    \small
    \includegraphics[width=1\linewidth]{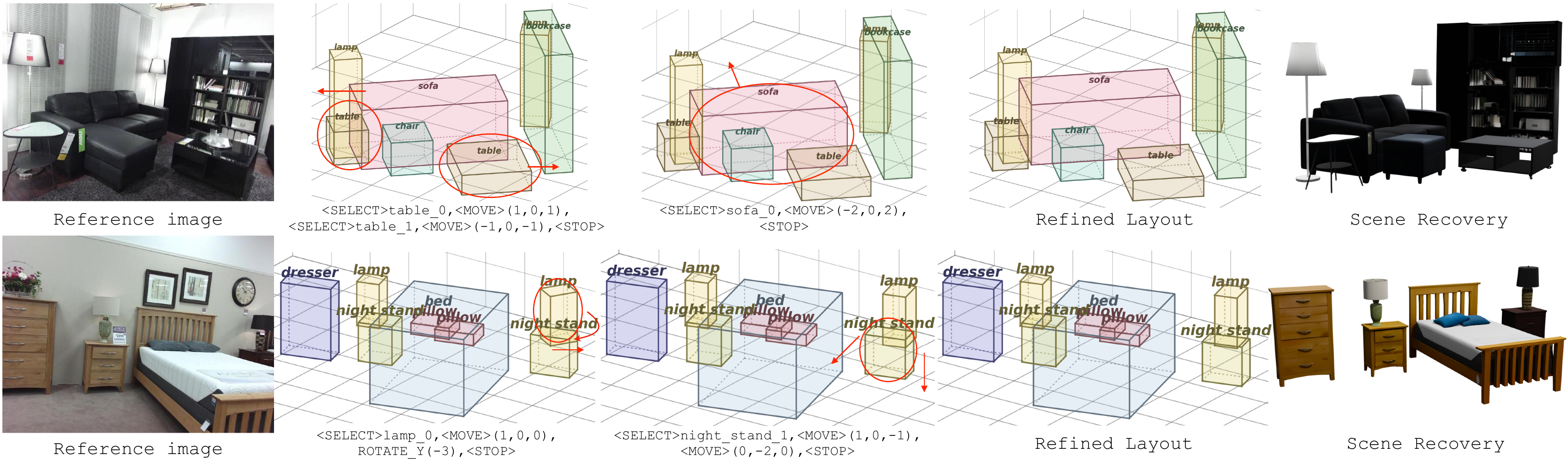}\\
    % \vspace{-0.05in}
    \caption{Example of our LaP guided action refinement process. We visualize each iteration's structured layout, together with the action sequences predicted by the LaP Planner.} 
    \vspace{-0.05in}
    \label{fig:refinement}
\end{figure}

\begin{figure}[t!]
    \centering
    \small
    \includegraphics[width=1\linewidth]{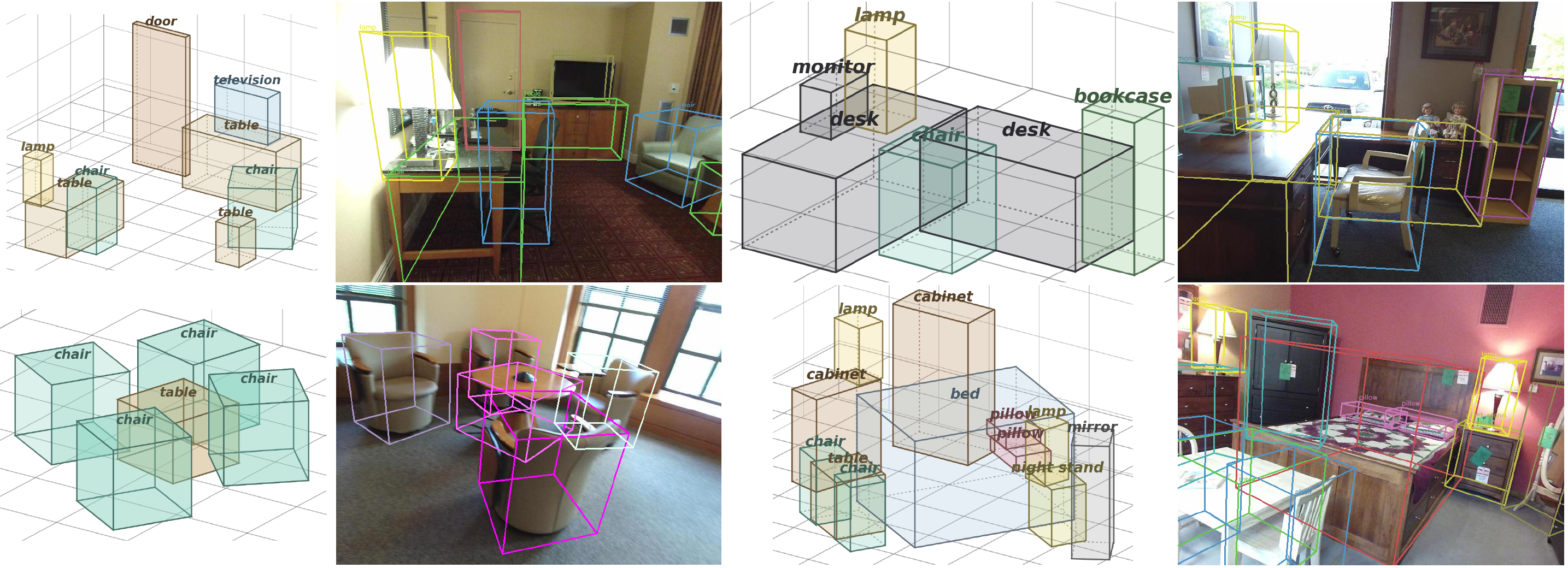}\\
    \vspace{-0.05in}
    \caption{More visualization of final 3D layout and reprojection. We back convert the structured layout back to the camera space and project the 3D boxes to the image for better visualization.} 
    \vspace{-0.05in}
    \label{fig:morevis}
\end{figure}

\begin{figure}[ht!]
    \centering
    \small
    %\vspace{-0.05in}
     \includegraphics[width=1\linewidth]{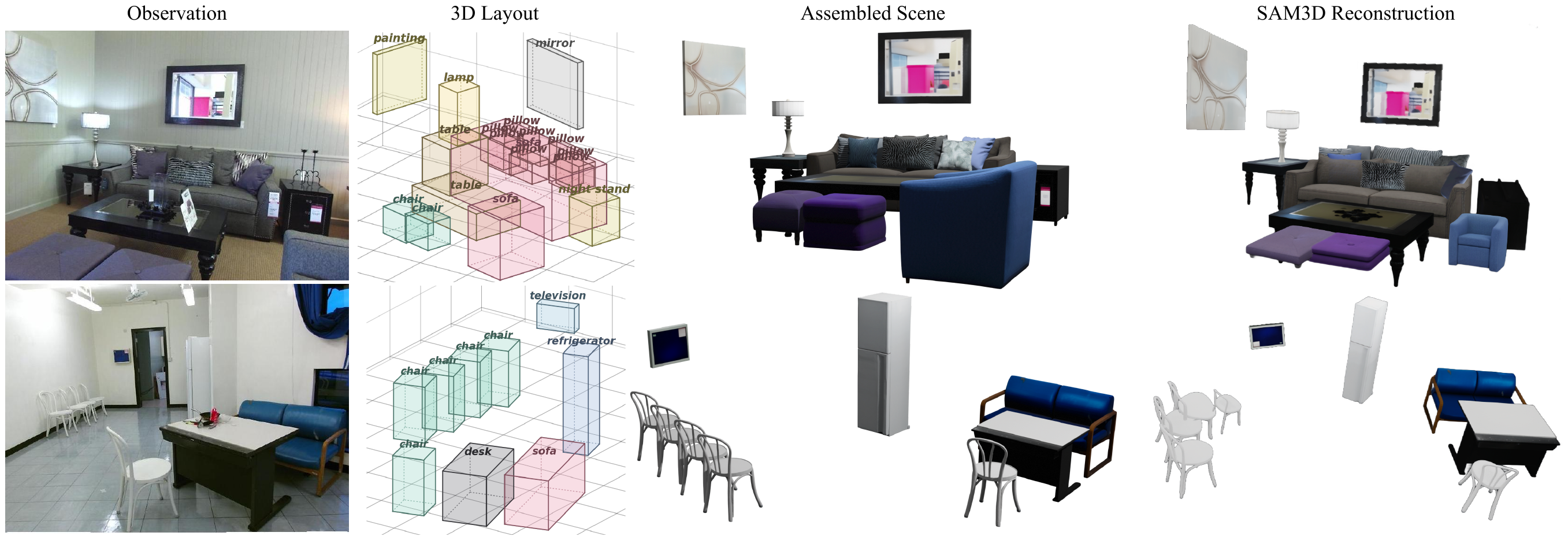}\\
    %\vspace{-0.05in}
    \caption{We show examples of our assembled scene with the estimated 3D layout. We compare with the 3D layout recovered from SAM3D~\cite{chen2025sam} using the same 3D assets for visual comparison.} 
    \vspace{-0.1in}
    \label{fig:assembly}
\end{figure}

\noindent\textbf{Qualitative Results.} Fig.~\ref{fig:perceiver} shows 3D grounding comparisons, where our Perceiver produces tighter alignment with object geometry than baselines, especially for objects with non-trivial orientations and partial occlusion. 
Fig.~\ref{fig:refinement} visualizes structured layouts before and after LaP Planner refinement (combined iterations for better visualization), progressively resolving physical inconsistencies such as collisions and floating objects, while preserving visual alignment.
We provide more visualizations of LaP Planner's refinement results 
across diverse scenes in Fig.~\ref{fig:morevis}.
Finally, Fig.~\ref{fig:assembly} presents 3D scenes reconstructed using 3D layout estimated from our perceive-then-plan pipeline, showing coherent object placements directly usable for downstream applications.

% \begin{wrapfigure}{r}{0.47\textwidth}
% \vspace{-.2in}
%   \begin{centering}
% \includegraphics[width=0.47\textwidth]{sources/languageedit.pdf}
%   \end{centering}
%   \vspace{-.2in}
%   \caption{\small Scene editing with instructions.}
%   \label{fig:editing}
%   % \vspace{-.2in}
% \end{wrapfigure}

\subsection{Ablation Study}
We ablate the key design choices of both components, with a 
particular focus on the LaP Planner to isolate the contribution 
of each element in our policy learning pipeline.
\begin{table}[t]
\centering
\vspace{.5em}

\caption{Ablation study on different designs in Perceiver grounding. We evaluate the impact of each component on 3D grounding quality. We denote the performance degradation in {\color{red}red} numbers. ``-'' denotes cumulative removal, the last row shows simply tuning without any additional components.}
\begin{tabular}{l cccc}
\toprule
Configuration & Reproj. IoU$\uparrow$ & Prec.{\tiny @(IoU=0.25)}$\uparrow$ & Prec.{\tiny @(IoU=0.5)}$\uparrow$ & Avg. DE$\downarrow$ \\
\midrule
Full Model (2B)             & 0.54 & 0.91 & 0.62 & 0.09 \\
\midrule
- Local-axis         & 0.41{\scriptsize\color{red}-0.13} & 0.79{\scriptsize\color{red}-0.12} & 0.48{\scriptsize\color{red}-0.14} & 0.10{\scriptsize\color{red}+0.01} \\
- Geo. Modulation         & 0.36{\scriptsize\color{red}-0.05} & 0.65{\scriptsize\color{red}-0.14} & 0.46{\scriptsize\color{red}-0.02} & 0.18{\scriptsize\color{red}+0.08} \\
\bottomrule
\end{tabular}
\vspace{-0.05in}
\label{tab:ablation_stage1}
\end{table}
For the Perceiver, we study the local-axis representation and 
geometry modulation in Tab.~\ref{tab:ablation_stage1}. Removing 
geometry modulation yields the largest drop across all metrics, 
confirming that fusing visual features with geometric cues is 
essential for accurate monocular 3D grounding.
\begin{table}[t!]
\centering

\caption{Ablation study on Stage 2 refinement. We evaluate the impact of each component on layout refinement quality and action numbers. Also, we denote the performance degradation in {\color{red}red}.}
\resizebox{\textwidth}{!}
{
\begin{tabular}{l ccccc}
\toprule
Configuration & Reproj. IoU$\uparrow$ & SVR\%$\downarrow$ & \# Collisions$\downarrow$ & Rot. Error$\downarrow$ &  Action Num.\\
\midrule
Full Model (2B)             & 0.65 & 4.85 & 1.46 & 9.05 & 3.67 \\
\midrule
w/o Grid-based Rep.         & 0.58{\scriptsize\color{red}-0.07} & 5.29{\scriptsize\color{red}+0.42} & 2.43{\scriptsize\color{red}+0.97} & 15.94{\scriptsize\color{red}+6.89}  & 3.53 \\
w/o SFT warm-up         & 0.56{\scriptsize\color{red}-0.09} & 5.12{\scriptsize\color{red}+0.27} & 2.18{\scriptsize\color{red}+0.72} & 13.43{\scriptsize\color{red}+4.38}  & 3.96 \\
w/o DPO         & 0.60{\scriptsize\color{red}-0.05} & 8.77{\scriptsize\color{red}+3.88} & 3.02{\scriptsize\color{red}+1.56} & 13.98{\scriptsize\color{red}+4.93}  & 4.73 \\
\bottomrule
\end{tabular}
}
\vspace{-0.05in}
\label{tab:ablation_stage2}
\end{table}
We then turn to the training pipeline of the LaP Planner. 
Tab.~\ref{tab:ablation_stage2} isolates each stage of our training 
recipe. Grid-based canonicalization transforms the continuous layout 
space into a tractable discrete action space, without which the 
Planner fails to produce coherent action sequences. SFT warm-up on 
expert trajectories yields partial refinement capability but plateaus 
before full convergence, action magnitudes are often over or 
under-estimated. Adding DPO enables the Planner to compare competing action sequences and deliver preferred sequences with minimal but sufficient actions in each iteration. 
% A central claim of our work is that layout refinement should be cast 
% as \emph{policy learning over discrete actions}, rather than direct 
% regeneration of corrected layouts. This 
% empirically validates our \textbf{Layout-as-Policy} formulation: 
% constraining refinement to discrete, object-level actions provides 
% a structural inductive bias that end-to-end regeneration lacks.

% \begin{figure}[ht!]
% \begin{minipage}[c]{0.51\linewidth}
%     \includegraphics[width=1.0\linewidth]{sources/languageedit.pdf}
%         \caption{\small Scene editing with language instructions.}
%     \label{fig:editing}
% \end{minipage}
% \hfill
% \begin{minipage}[c]{0.47\linewidth}
%     \includegraphics[width=\linewidth]{sources/instructional_refinement.pdf}
%   \caption{\small An example of instructional refinement.}
%   \label{fig:instructional_refine}
% \end{minipage}
% \vspace{-.2in}
% \end{figure}

% \subsection{Downstream applications}
% Our structured action-level interface naturally extends the LaP Planner to downstream tasks. Fig.~\ref{fig:editing} illustrates language-guided editing, where user instructions are translated into action sequences that modify the layout while preserving plausibility, we demonstrate the edited layout and recovered scene rendering together. The editing reuse the same action vocabulary, showcasing the extensibility of our perceive-then-plan framework.

\subsection{Generalization \& Comparison with Non-VLM Methods}
\label{sec:generalization}
% Tables 5 and 6 share one float to save vertical space
\begin{table}[t]
\centering
\footnotesize
\setlength{\tabcolsep}{5pt}
\caption{Zero-shot results on unseen datasets. Gain over Perceiver initialization are marked in {\color{red}red}:.}
\label{tab:crossdataset}
\begin{tabular}{l|l cccc}
\toprule
Dataset & Method & Reproj. IoU$\uparrow$ & SVR (\%)$\downarrow$ & \# Collisions$\downarrow$ & Rot. Err.$\downarrow$ \\
\midrule
\multirow{4}{*}{3D-FRONT}
 & N3D-VLM         & 0.31 & 10.11 & 3.38 & 44.54 \\
 & DetAny3D~\cite{zhang2025detany3d}  & 0.55 & 17.65 & 1.61 & 24.16 \\
 & Ours (Perceiver)                   & 0.57 & 7.47  & 2.41 & 14.50 \\
 & Ours (Full)  & \textbf{0.70}{\scriptsize\color{red}+0.13} & \textbf{2.83}{\scriptsize\color{red}-4.64} & \textbf{0.94}{\scriptsize\color{red}-1.47} & \textbf{8.82}{\scriptsize\color{red}-5.68} \\
\midrule
\multirow{4}{*}{ARKitScenes}
 & N3D-VLM         & 0.23 & 32.53 & 2.70 & 45.74 \\
 & DetAny3D~\cite{zhang2025detany3d}  & 0.44 & 26.52 & 3.28 & 33.66 \\
 & Ours (Perceiver)                   & 0.42 & 20.80 & 3.02 & 26.43 \\
 & Ours (Full)  & \textbf{0.50}{\scriptsize\color{red}+0.08} & \textbf{12.44}{\scriptsize\color{red}-8.36} & \textbf{1.78}{\scriptsize\color{red}-1.24} & \textbf{20.05}{\scriptsize\color{red}-6.32} \\
\bottomrule
\end{tabular}
\vspace{.2in}
\centering
\footnotesize
\setlength{\tabcolsep}{4pt}
\caption{Comparison with non-VLM methods on SUN RGB-D, using the same input 2D boxes.}
\label{tab:nonvlm}
\begin{tabular}{l cccccc}
\toprule
Method & Reproj. IoU$\uparrow$ & Prec.{\tiny@(IoU=0.25)}$\uparrow$ & Prec.{\tiny@(IoU=0.5)}$\uparrow$ & Avg. DE$\downarrow$ & SVR (\%)$\downarrow$ & \# Coll.$\downarrow$ \\
\midrule
DetAny3D~\cite{zhang2025detany3d} & 0.50 & 0.81 & 0.59 & 0.16 & 21.91 & 2.70 \\
OVMono3D~\cite{yao2024ovmono3d}   & 0.57 & \textbf{0.92} & 0.60 & 0.15 & 15.13 & 3.04 \\
SAM 3D         & 0.41 & 0.84 & 0.51 & 0.12 & 6.62 & 2.58 \\
\midrule
Ours (Perceiver)                  & 0.54 & 0.91 & 0.62 & 0.09 & 11.63 & 3.62 \\
Ours (Full)                       & \textbf{0.65} & 0.91 & \textbf{0.63} & \textbf{0.07} & \textbf{4.85} & \textbf{1.46} \\
\bottomrule
\end{tabular}
\vspace{-.2in}
\end{table}

\noindent\textbf{Zero-shot cross-dataset evaluation.} We evaluate our pipeline without any dataset-specific training on 3D-FRONT~\cite{fu20213dfront} and ARKitScenes~\cite{baruch2021arkitscenes}. As shown in Tab.~\ref{tab:crossdataset}, the LaP Planner consistently improves its Perceiver initialization on both unseen datasets, and the full pipeline outperforms N3D-VLM~\cite{wang2025n3d} and DetAny3D~\cite{zhang2025detany3d} (grounding results in Appendix~\ref{app:extra}).

\noindent\textbf{Comparison with non-VLM methods.} Tab.~\ref{tab:nonvlm} compares with monocular 3D detectors (DetAny3D~\cite{zhang2025detany3d}, OVMono3D~\cite{yao2024ovmono3d}) and the image-to-3D method SAM 3D~\cite{chen2025sam}, whose meshes are fitted with oriented boxes and placed by its predicted 6-DoF poses. Our method is best on five of six metrics, with OVMono3D marginally higher on Prec.@(IoU=0.25).

\subsection{Downstream applications}
\begin{wrapfigure}{r}{0.47\textwidth}
\vspace{-.2in}
  \begin{centering}
\includegraphics[width=0.47\textwidth]{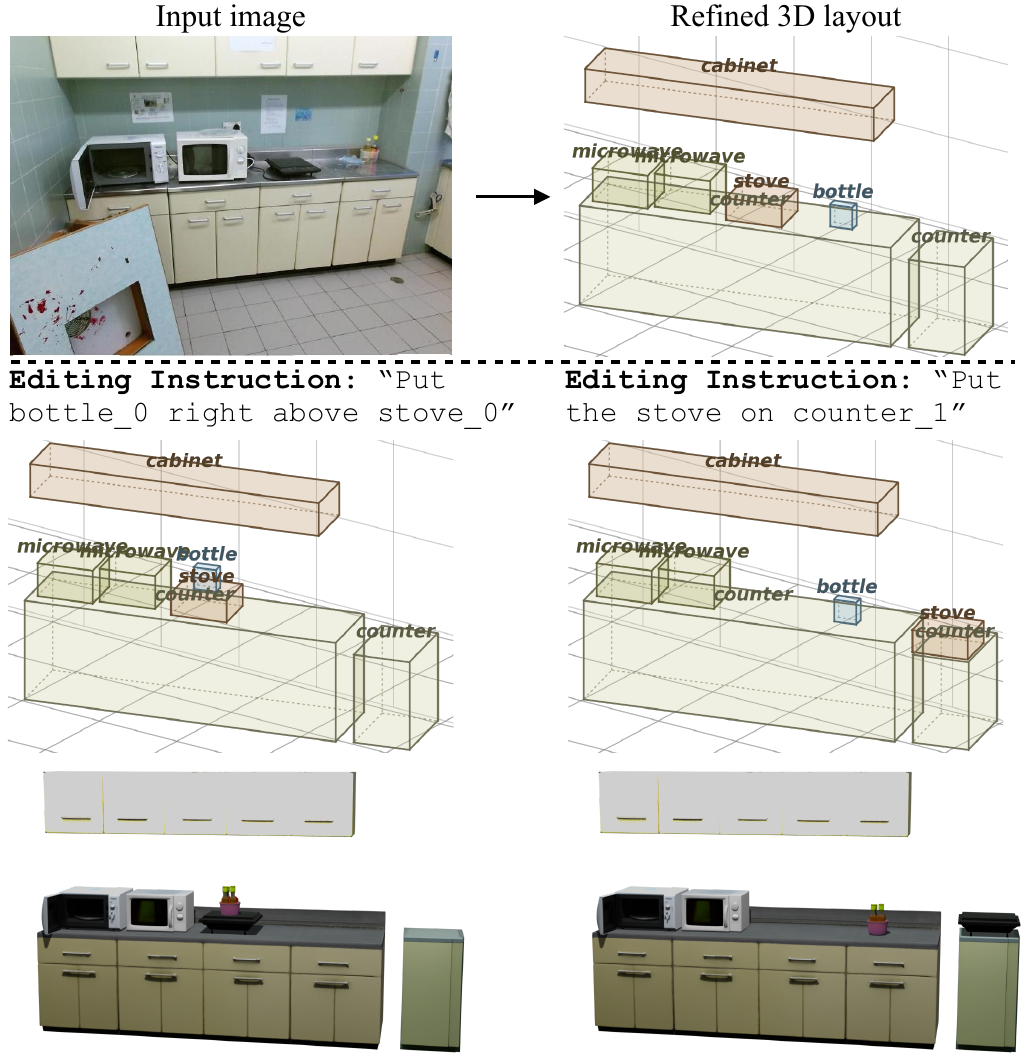}
  \end{centering}
  \vspace{-.2in}
  \caption{\small Scene editing with instructions.}
  \label{fig:editing}
  \vspace{-.2in}
\end{wrapfigure}
Our structured action interface naturally extends the LaP Planner to downstream tasks. Fig.~\ref{fig:editing} illustrates language-guided editing, where user instructions are translated into action sequences from the same vocabulary, modifying the layout while preserving plausibility.
% \vspace{-.1in}
% \section{Conclusion}
% \label{sec:conclu}
% \vspace{-.1in}
% In this work, we propose a perceive-then-plan 
% framework for monocular 3D layout estimation, and introduced \textbf{Layout-as-Policy (LaP)} that recasts the task as a sequential 
% decision-making problem. 
% %
% Our approach decomposes the task into two 
% specialized VLMs: a geometry-enhanced \textbf{Perceiver} that grounds 
% 3D objects from an input image, and a \textbf{LaP Planner} that 
% iteratively refines the layout through discrete, policy-driven actions, 
% trained via supervised warm-up followed by direct preference 
% optimization. 
% %
% Experiments demonstrate substantial improvements in both 3D grounding accuracy and physical plausibility over strong VLM baselines, while naturally supporting 
% downstream applications such as language-guided editing and interactive 
% 3D grounding.
% %
% Opening up new possibilities for Vision-Language Models in 3D reasoning and real-world scenarios.
% %
% {While our framework achieves strong results, it relies on 
% 2D object detections as input, and errors or missed detections 
% may propagate through the pipeline, which remains a limitation 
% of the current work.}

\vspace{-.1in}
\section{Conclusion}
\label{sec:conclu}
\vspace{-.1in}
In this work, we propose a perceive-then-plan framework for monocular 3D layout estimation, and introduce \textbf{Layout-as-Policy (LaP)} that recasts the task as a sequential decision-making problem. 
Our approach decomposes the task into two specialized VLMs: a geometry-enhanced \textbf{Perceiver} that grounds 3D objects from an input image, and a \textbf{LaP Planner} that iteratively refines the layout through discrete actions in a closed loop, trained via supervised warm-up followed by direct preference optimization. 
Experiments show improved 3D grounding accuracy and physical plausibility, generalization to unseen datasets, and natural support for downstream applications such as scene editing.

\noindent\textbf{Limitations.} Noisy or missing 2D detections may propagate errors (Appendix~\ref{app:extra}); training excludes scenes with over 30 objects; and the SVD-based gravity estimation may be misaligned when the floor is barely visible. Multi-view input is left for future work.

\newpage
{
    \small
    \bibliographystyle{ieeenat_fullname}
    \bibliography{main}
}

%%%%%%%%%%%%%%%%%%%%%%%%%%%%%%%%%%%%%%%%%%%%%%%%%%%%%%%%%%%%
\newpage
\appendix
% \clearpage
% \setcounter{page}{1}
% \setcounter{section}{0}
% \renewcommand\thesection{\Alph{section}}
% \counterwithin{figure}{section}
% \counterwithin{table}{section}
% \renewcommand\thefigure{\thesection.\arabic{figure}}
% \renewcommand\thetable{\thesection.\arabic{table}}
\section{More Implementation Details}
\label{app:imple}
\subsection{Model Architecture and Training Details}

Both the Perceiver and the LaP Planner are built on Qwen3-VL-2B-Instruct and Qwen3-VL-8B-Instruct~\cite{bai2025qwen3}.
For the Perceiver, we use VGGT-1B~\cite{wang2025vggt} as an external geometry encoder. Geometry features are injected into the visual tokens through the gated FiLM modulation in Eq. (3). The vision encoder, its projector and the geometry encoder are kept frozen; the language model is fully fine-tuned together with the geometry merger and the modulation module.
We fine-tune the LaP Planner using LoRA adapters~\cite{hu2022lora} with rank $r = 16$ applied to the attention projection matrices ($W_q$, $W_k$, $W_v$, $W_o$). We set $\alpha = 32$ and apply a dropout rate of 0.05. The DPO temperature parameter is set to $\beta = 0.1$ for stable optimization.
The Perceiver is trained on 8 NVIDIA RTX Pro 6000 96G GPUs using AdamW with a learning rate of $1\times10^{-4}$, cosine scheduling, and batch size 16. It is trained for 2 epochs, taking approximately 14 hours for the 2B model. The memory usage varies according to the scene complexity, but no larger than 90 GB.
For the LaP Planner, we first perform SFT warm-up using 8 NVIDIA RTX 6000 48G GPUs with AdamW, a learning rate of $5\times10^{-5}$, and 1 training epoch. We then initialize DPO training from the SFT checkpoint and train for 2 additional epochs with a learning rate of $1\times10^{-6}$. The SFT and DPO stages take approximately 10 hours and 13 hours, respectively, for the 2B model. The memory usage is often 42 to 46GB.

\subsection{LaP Planner Training Set}
\label{app:laptraining}
In both SFT and DPO stages, we construct training data using a mixture of three types of supervision. We include 20\% samples initialized directly from ground-truth 3D layouts without perturbation. For these samples, the action sequence consists solely of a \texttt{<STOP>} token, which teaches the model to terminate refinement when a satisfactory layout is reached.
The remaining data consists of 30\% Perceiver-initialized layouts and 50\% synthetically perturbed layouts, improving robustness to imperfect initialization and increasing data diversity.
For DPO training, we further construct preference pairs using a metric-guided filtering strategy. We first align the ground plane of initialized layouts with the ground-truth scene to reduce global translation ambiguity. For each Perceiver-initialized sample, we generate candidate action sequences and evaluate the resulting intermediate layouts using geometric and structural metrics.
We track object identities involved in each action sequence and construct comparisons conditioned on identical object subsets. A trajectory is selected as the preferred one only if it consistently yields better improvement according to layout quality metrics. Samples without a clear dominance relationship are discarded. This filtering strategy produces high-quality preference pairs, leading to more stable DPO optimization and improved training robustness.
It is worth mentioning that, the perturbation dataset generation leads to the increase in different types of physical inconsistencies with our perturbation actions, e.g., collisions, wrong support relations, which can naturally enforce the model to learn both visual alignment and physical plausibility jointly in  the same training phase. 

\subsection{Evaluation Setting}
\label{app:eval}
\paragraph{Evaluation metrics.} 
We evaluate the Perceiver's 3D grounding quality with four metrics:
\begin{itemize}
\item \textbf{Reprojection IoU (Reproj. IoU $\uparrow$).} 
We project the predicted 3D box onto the image plane and compute the 2D IoU between the box enclosing the projected corners and the visible 2D box of the object. Projection uses a pinhole camera with focal length equal to the image height and the principal point at the image center.

\item \textbf{Precision at IoU=0.25 (Prec.@(IoU=0.25) $\uparrow$).} 
The fraction of predicted boxes whose Reprojection IoU with the 
matched ground-truth box exceeds 0.25. This corresponds to a 
loose-alignment regime, capturing whether the model identifies 
the correct rough region of each object.

\item \textbf{Precision at IoU=0.50 (Prec.@(IoU=0.5) $\uparrow$).} 
The fraction of predictions exceeding the stricter threshold of 
0.50. Together with Prec.@(IoU=0.25), this provides a 
coarse-to-fine view of grounding accuracy.

\item \textbf{Average Depth Error (Avg. DE $\downarrow$).} 
The mean relative error $|\hat{z} - z| / z$ between predicted and ground-truth box centers along the camera viewing direction, averaged 
over all matched objects. While Reprojection IoU evaluates 
in-image alignment, Depth Error specifically isolates the model's 
ability to recover absolute 3D position, which is the most 
challenging aspect of monocular 3D grounding.
\end{itemize}

We evaluate the LaP Planner along two complementary axes: visual 
alignment and physical plausibility using four metrics:

\begin{itemize}
\item \textbf{Reprojection IoU (Reproj. IoU $\uparrow$).} 
Same as in Perceiver evaluation. Reported here to verify that 
iterative refinement does not drift away from image evidence.

\item \textbf{Support Violation Rate (SVR $\downarrow$).} 
The fraction of objects in a scene that lack a valid support relation in the canonicalized layout. Operating in grid coordinates, an object is considered supported if either a) its bottom is within 1 grid unit of the ground plane (y=0), or b) its bottom is within 1 grid unit of another object's top with at least 50\% horizontal overlap in the XZ plane; otherwise it is counted as a violation. Wall-mounted classes (paintings, mirrors, boards, clocks) are excluded as horizontal-surface support is ill-defined for them, and small objects (cups, bottles, books, towels) are excluded as their support relations are sensitive to annotation noise at our grid resolution. SVR directly measures one of the most prevalent failure modes in VLM-based grounding: objects floating in mid-air or detached from their supporters.

\item \textbf{Collision Count (\# Collisions $\downarrow$).} 
The number of pairwise 3D bounding box penetrations within a scene, 
where two boxes are considered colliding if their intersection 
volume exceeds 20\% of the smaller box's volume. This metric captures the second major class of physical inconsistency: objects penetrating each other. Since there are also collisions in the ground truth data, we do not take those into account when computing Collisions for the results.

\item \textbf{Rotation Error (Rot. Err. $\downarrow$).} 
The mean angular difference (in degrees) between predicted and ground-truth object orientations around the gravity axis, averaged over all matched objects. Angular differences are computed as the minimum rotation modulo 360°, i.e., wrapped to [0°,180°]. We exclude object classes that exhibit rotational symmetry around the gravity axis, as their orientation is ill-defined; lamps, round tables, and stools are representative examples.
\end{itemize}

\subsection{Implementation Details of methods compared with LaP Planner.} 
\label{app:complap}
For GPT-4o, we simply use the same prompt and iterative refinement as our LaP Planner. 
For the two comparison methods applied in Tab.~\ref{tab:stage2_refinement}, we introduce the implementation details.

1) Rule-Based Refinement: The rule-based baseline applies a fixed sequence of geometric corrections in the discretized grid space, requiring no learned model or image input. It runs two alternating passes based on the contact scene-graph generated: a de-floating pass that snaps each floating object onto its corresponding supporter or lowers it to the ground, and a de-collision pass that separates all axis-aligned bounding-box collisions by pushing objects apart along the minimum-penetration axis, preferring horizontal displacement. Object rotation and size are left unchanged.

2) One-shot VLM direct refinement: The direct prediction baseline prompts the finetuned Qwen3-VL-8B-Instruct model with the same perturbation and gt 3D layout data. Given the RGB image and a structured text description of the perturbed layout (object category, 2D bounding box, grid position, size, and yaw), the model is asked to output a corrected JSON layout in a single pass. This baseline tests whether a VLM can perform layout refinement from instruction alone, without the action-space supervision that LaP Planner employs.

We also test the \textbf{multi-rounds} performance under this setting (\textit{One-shot VLM direct refinement}). With 3 rounds of refinement, the Reproj. IoU drops to 0.56, and SVR, Collisions, Rot. Err, and Avg. DE remains approximately unchanged or degraded (\textbf{10.12}, \textbf{3.43}, \textbf{15.05}, and \textbf{0.11}), further showcasing direct prediction refinement can hinder the Vision-Language Models from understanding the scene state, which does not produce layout that are marginally better.

\subsection{Action Sequence Degradation}
\label{app:actionseq}
To construct rejected samples ($y^-$) in the perturbed data synthesis part for DPO training, we apply a 
suite of degradation operations to the ground-truth corrective 
action sequence ($y^+$). Each rejected sample is obtained by 
applying 1--2 randomly chosen degradations to the GT sequence, 
ensuring that rejected samples remain 
suboptimal or collapsed.

\noindent \textbf{Numerical shift.} 
We perturb the parameters of transform actions by small offsets, 
e.g., \texttt{MOVE($dx$, $dy$, $dz$)} $\to$ \texttt{MOVE($dx + \delta_x$, $dy + \delta_y$, $dz + \delta_z$)}, 
where $\delta_x, \delta_y, \delta_z \sim \mathcal{U}\{-3, -2, -1, 1, 2, 3\}$ in grid 
units. Similar shifts are applied to 
\texttt{ROTATE\_Y} and \texttt{RESIZE} parameters. We additionally 
introduce \emph{magnitude-only} variants that scale the action 
parameters by a factor sampled from $\mathcal{U}(0.3, 0.7)$ 
(undershoot) or $\mathcal{U}(1.5, 2.5)$ (overshoot), preserving the 
correct direction but with incorrect magnitude. Both forms produce 
actions that point in roughly the right direction but with imprecise 
magnitudes, teaching the Planner to favor precise parameter values.

\noindent \textbf{Nuisance actions.} 
We insert redundant or no-op actions into the sequence, such as 
duplicate \texttt{SELECT} tokens for the same object or transform 
actions whose effect cancels out (e.g., \texttt{MOVE(2, 0, 0)} 
followed by \texttt{MOVE(-2, 0, 0)}). We also include 
\emph{over-correction} variants where the GT trajectory is augmented 
with extra action blocks targeting objects that were not modified 
during perturbation. These bloat the sequence without contributing 
to refinement and discourage verbose outputs that touch already 
correct objects.

\noindent \textbf{Missing actions.} 
We randomly drop a subset of actions from the GT sequence, leaving 
the corresponding object's perturbation uncorrected. In the most 
extreme case, the entire trajectory is replaced by a single 
\texttt{STOP} token (\emph{premature termination}), simulating the 
failure mode where the Planner declares convergence despite visible 
errors. This category teaches the Planner to prefer complete 
refinement over partial fixes and to avoid early termination, 
particularly important for scenes with multiple perturbed objects.

\noindent \textbf{Wrong action types.} 
We replace actions with an incompatible type, e.g., 
\texttt{MOVE($dx$, $dy$, $dz$)} $\to$ \texttt{ROTATE\_Y($\theta$)}, 
where the new action does not address the error in the 
layout. We also include \emph{direction-flip} variants where 
\texttt{delta} signs are negated 
(e.g., \texttt{MOVE($dx, dy, dz$)} $\to$ \texttt{MOVE($-dx, -dy, -dz$)}), 
producing actions that move objects in the opposite direction. 
This make the Planner to diagnose what kind of 
correction each object requires and in which direction.

\subsection{Prompting Format}
We provide the full prompting templates used during training and 
inference for both the Perceiver and the LaP Planner. Placeholders 
in curly braces (e.g., \texttt{\{image\}}, \texttt{\{layout\}}) are 
replaced with task-specific content at runtime.

\paragraph{Perceiver prompt.} 
The Perceiver is prompted to output 3D bounding boxes in the 
canonicalized gravity-aligned coordinate system, given an image and 
a list of 2D detections.

\begin{tcolorbox}[
  title=Perceiver,
  label=prompt:perceiver,
  colback=gray!5,
  colframe=gray!50!black,
  fonttitle=\bfseries,
  breakable
]
\small
\texttt{Here are the detected 2D bounding boxes in this image, format:
(id, class, x1, y1, x2, y2): (1, lamp, 841.0000, 378.0000, 917.0000,
555.0000)}\\
\texttt{(2, pillow, 568.0000, 435.0000, 778.0000, 526.0000)}\\
\texttt{(3, bed, 449.0000, 287.0000, 923.0000, 891.0000)}\\
\texttt{$\ldots$\,(\textit{2D bounding box list for each object})}\\[3pt]
\texttt{Output a json list, where each entry is a 3D bounding box in
the CAMERA coordinate system that corresponds to a given 2D bounding
box. Each 3D bounding box must follow exactly this schema:}\\[3pt]
\texttt{\{"id": int,}\\
\texttt{\quad"class": str,}\\
\texttt{\quad"center": [position\_u, position\_v, position\_x],}\\
\texttt{\quad"size": [scale\_x, scale\_y, scale\_z],}\\
\texttt{\quad"x\_axis": [x1, x2, x3],}\\
\texttt{\quad"y\_axis": [y1, y2, y3]\}}\\[3pt]
\texttt{Rules: The coordinate system is CAMERA: +u right, +v down,
+x away from camera.}\\
\texttt{"center" is the 3D box center in the GLOBAL camera
coordinates.}\\
\texttt{"size" is the length of the three edges along the box axes.}\\
\texttt{"x\_axis" and "y\_axis" are unit vectors defining the LOCAL
object frame of the box, must be orthogonal.}\\
\texttt{Do NOT include explanations or extra text.}\\
\texttt{All numbers must be valid floating point values.}
\end{tcolorbox}

\paragraph{LaP Planner prompt.} 
The LaP Planner is prompted to emit a sequence of corrective action sequences 
that refine a given initial layout with a system prompt and a user instruction.

\begin{tcolorbox}[
  title=Planner System Prompt,
  label=prompt:planner_system,
  colback=gray!5,
  colframe=gray!50!black,
  fonttitle=\bfseries,
  breakable
]
\small
\texttt{You are a 3D layout refinement agent. Given an image and the
current 3D layout of detected objects, your task is to correct errors
in object positions, orientations, and sizes. If the layout is already
correct, output STOP immediately.}\\[3pt]
\textbf{\texttt{\#\# Scene Representation}}\\[2pt]
\texttt{Each object in the scene is described with:}\\
\texttt{\quad - obj\_id and category: object identity}\\
\texttt{\quad - bbox: [x1, y1, x2, y2] \# 2D bounding box in
normalized image pixel coordinates (0--1000)}\\
\texttt{\quad - pos: [gx, gy, gz] \# 3D position in grid units
(1 grid unit = 10\,cm), where gy is the bottom of the object}\\
\texttt{\quad - size: [gw, gh, gl] \# width, height, length in
grid units}\\
\texttt{\quad - yaw: orientation index (0--23, each step =
15\textdegree)}\\[3pt]
\textbf{\texttt{\#\# Coordinate Axes}}\\[2pt]
\texttt{\quad - X: horizontal, increases toward image right}\\
\texttt{\quad - Y: vertical, increases upward}\\
\texttt{\quad - Z: depth, increases forward into the scene}\\[3pt]
\textbf{\texttt{\#\# Action Space}}\\[2pt]
\texttt{SELECT obj\_N \quad \# choose target object}\\
\texttt{MOVE [dx, dy, dz] \quad \# adjust position}\\
\texttt{ROTATE\_Y [d] \quad\quad\enspace \# adjust orientation (each
unit = 15\textdegree)}\\
\texttt{RESIZE [d] \quad\quad\enspace \# uniformly adjust size}\\
\texttt{STOP \quad\quad\quad\quad\enspace \# end the sequence}\\[3pt]
\textbf{\texttt{\#\# Rules}}\\[2pt]
\texttt{\quad - Output ONLY actions, one per line, integers only.}\\
\texttt{\quad - SELECT before correcting an object.}\\
\texttt{\quad - If the layout already looks correct, output STOP
immediately.}\\
\texttt{\quad - Fix the most significant errors first.}\\
\texttt{\quad - Prefer fewer actions. Stop when no further correction
is needed.}
\end{tcolorbox}

\begin{tcolorbox}[
  title=Planner,
  label=prompt:planner_instruction,
  colback=gray!5,
  colframe=gray!50!black,
  fonttitle=\bfseries,
  breakable
]
\small
\texttt{<image>}\\[3pt]
\texttt{Examine the image and the detected 3D layout below. Identify
and correct any errors in object positions, orientations, or sizes.}\\[3pt]
\textbf{\texttt{Current scene layout:}}\\[2pt]
\textbf{\texttt{\#\# Scene Layout (grid-based, 1 unit = 10\,cm)}}\\[2pt]
\texttt{obj\_0 cabinet}\\
\texttt{\quad bbox: [\,\ldots\,]}\\
\texttt{\quad pos: [\,\ldots\,]}\\
\texttt{\quad size: [\,\ldots\,]}\\
\texttt{\quad yaw: \ldots}\\
\texttt{\ldots\,(\textit{3D layout list for all objects})}
\end{tcolorbox}

\paragraph{Contact graph prompt.} 
For scene assembly, the LaP Planner (with LoRA 
disabled) is prompted to extract pairwise support relations.

\begin{tcolorbox}[
  title=Contact Graph Prompt,
  label=prompt:contact,
  colback=gray!5,
  colframe=gray!50!black,
  fonttitle=\bfseries,
  breakable
]
\small
\texttt{Given an indoor image and a list of detected 2D bounding
boxes, identify the physical contact relation for each object.
For every detected object, assign exactly one of the following
contact types:}\\[3pt]
\texttt{\quad - FLOOR: the object rests directly on the floor}\\
\texttt{\quad - ON obj\_id: the object rests on top of another
detected object}\\
\texttt{\quad - FREE: the object has no contact with the floor or
any other object (e.g., wall-mounted, hanging, floating)}\\[3pt]
\texttt{For each detected object, output a single line:}\\[3pt]
\texttt{\quad <CONTACT> id: \{id\} class: \{class\} relation:
\{FLOOR $|$ ON obj\_id $|$ FREE\} </CONTACT>}\\[3pt]
\texttt{Examples:}\\
\texttt{\quad <CONTACT> id: 3 class: bed relation: FLOOR
</CONTACT>}\\
\texttt{\quad <CONTACT> id: 2 class: pillow relation: ON 3
</CONTACT>}\\
\texttt{\quad <CONTACT> id: 1 class: lamp relation: ON 5
</CONTACT>}\\[3pt]
\texttt{Input image:} \texttt{\{image\}}\\[2pt]
\texttt{Detections (id, class, x1, y1, x2, y2):}\\
\texttt{(1, lamp, 841.0000, 378.0000, 917.0000, 555.0000)}\\
\texttt{(2, pillow, 568.0000, 435.0000, 778.0000, 526.0000)}\\
\texttt{(3, bed, 449.0000, 287.0000, 923.0000, 891.0000)}\\
\texttt{$\ldots$\,(\textit{full detection list})}\\[3pt]
\texttt{Output one <CONTACT>...</CONTACT> line per detected
object. Do NOT skip any object.}
\end{tcolorbox}

\begin{wrapfigure}{r}{0.5\textwidth}
  \begin{centering}
\includegraphics[width=0.5\textwidth]{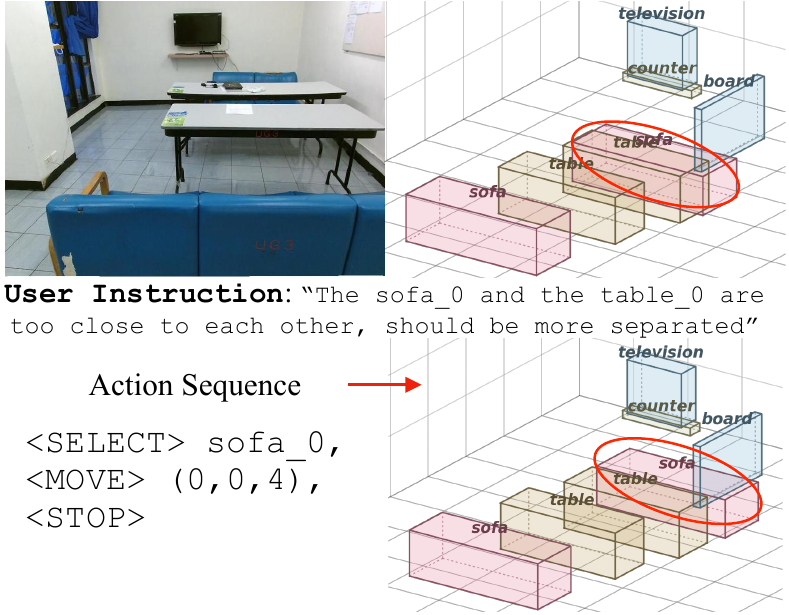}
  \end{centering}
  \vspace{-.2in}
  \caption{Another downstream application of instructional refinement using user input language.}
  \label{fig:editing2}
  \vspace{-.4in}
\end{wrapfigure}

\subsection{Scene Assembly}
After layout refinement by the LaP Planner, we populate the 
predicted 3D boxes with actual 3D assets (retrieved from existing 
datasets or generated by 3D generative models, we generate with SAM 3D in the main paper for better visualization) and apply a 
gravity simulation step to resolve residual placement artifacts 
caused by asset-box geometry mismatches.

\paragraph{Contact graph generation.} 
We disable the LoRA adapter on the LaP Planner and prompt it 
(Box~\ref{prompt:contact}) to generate a contact scene graph 
$\mathcal{G}_{\text{contact}}$ from the input image. Each edge in 
$\mathcal{G}_{\text{contact}}$ encodes a directed 
\texttt{supported-by} relation between two objects (e.g., 
\texttt{pillow supported by bed, bed supported by ground}...). 
Disabling LoRA ensures we use 
the backbone's general visual reasoning ability without bias toward 
action-token generation.

\paragraph{Relational bundle construction.}
We traverse $\mathcal{G}_{\text{contact}}$ to partition objects into
three groups according to their assigned contact relation. Objects
labeled \texttt{FLOOR} are treated as ground-level anchors. Objects
labeled \texttt{ON}~$obj_i$ are attached to their supporter $obj_i$,
forming a \emph{relational bundle}: a rigid composite of a supporter
and all objects transitively stacked upon it. Objects labeled
\texttt{FREE} (e.g., wall-mounted lamps) are
excluded from gravity settling and kept at their refined positions
throughout simulation. Treating bundles as rigid composites prevents
supporters and supportees from drifting apart during gravity settling,
preserving the relational structure established during layout
refinement.

\paragraph{Gravity simulation.}
We run a single-pass rigid-body simulation to drive
the refined layout toward static equilibrium. The floor is fixed at
$y = 0$ with gravity. \texttt{FLOOR} objects
and relational bundles are initialized with a small clearance gap
along the gravity axis to avoid initial interpenetrations.

\section{More Results}
\label{supp:morevis}
We provide additional qualitative results to further validate the 
effectiveness of our proposed framework across different components 
and applications. Since Hypersim is more of a simulation dataset, we present more results to demonstrate the ability in 3D layout estimation from monocular observations.

\subsection{Downstream Application}
We showcase an additional downstream application, i.e., language-instructed refinement in Fig.~\ref{fig:editing2}, showing that with vision-language models, our framework support better extendability.

\begin{figure}[ht!]
    \centering
    \small
    \includegraphics[width=1\linewidth]{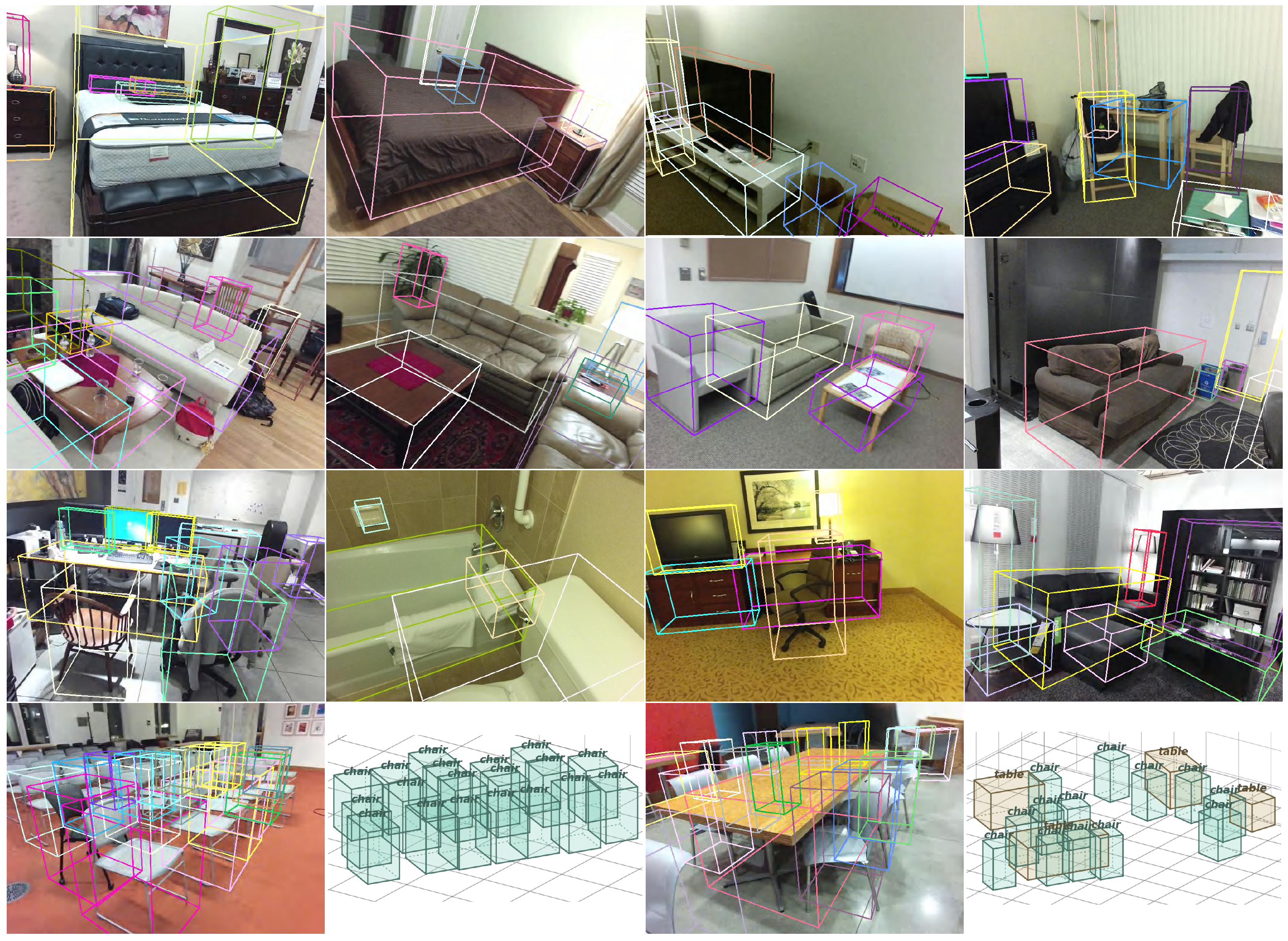}\\
    \vspace{-0.1in}
    \caption{We present more visualization results of our Perceiver on real-world captures.} 
    \vspace{-0.1in}
    \label{fig:supp_perceiver}
\end{figure}

\begin{figure}[ht!]
    \centering
    \small
    \includegraphics[width=1\linewidth]{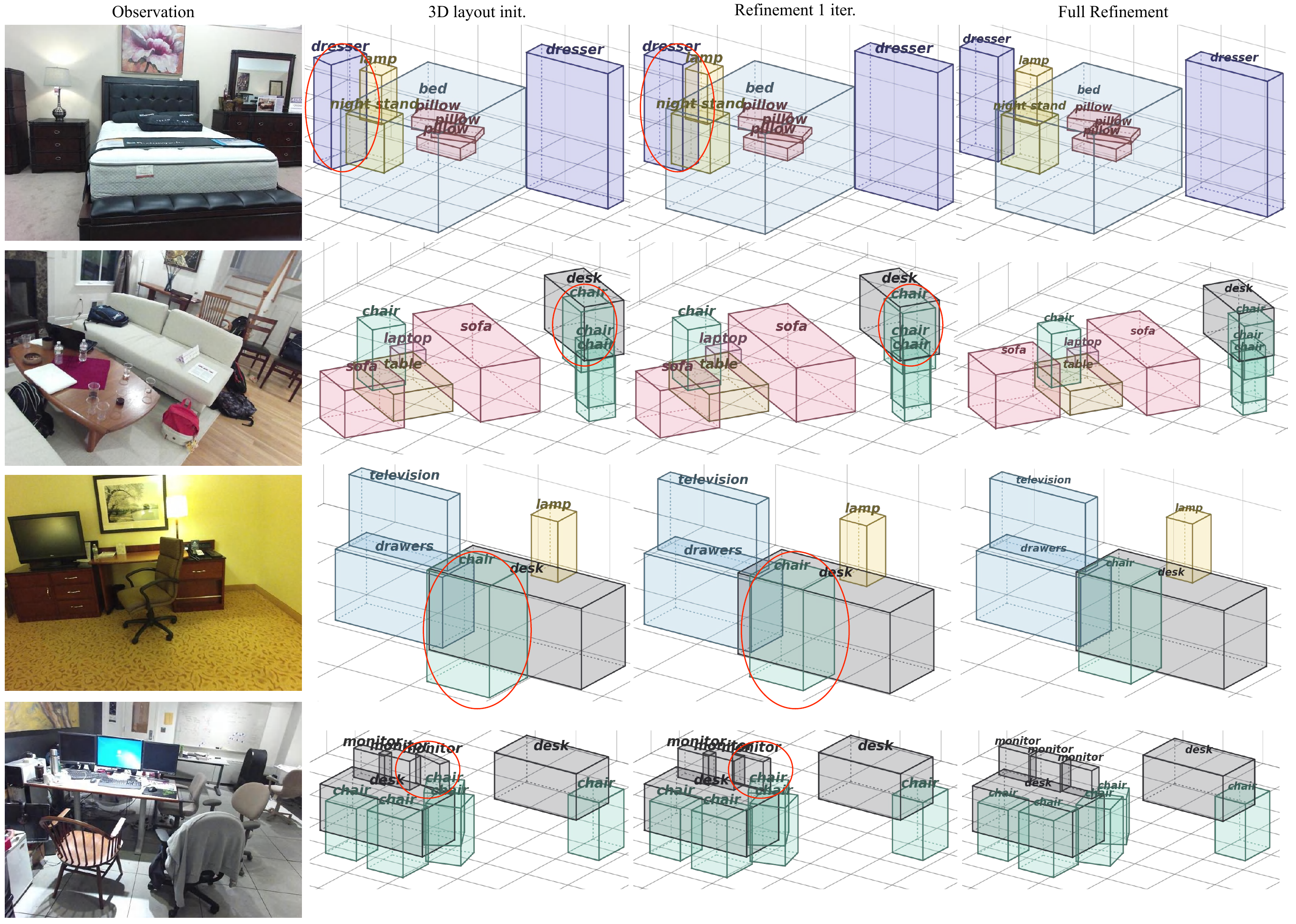}\\
    \vspace{-0.1in}
    \caption{We present more visualization results of our LaP Planner.} 
    \vspace{-0.1in}
    \label{fig:supp_lapplanner}
\end{figure}

\subsection{3D Grounding}
\label{3dgrounding}
We present additional 3D grounding results in Fig.~\ref{fig:supp_perceiver}. 
These visualizations demonstrate the Perceiver's ability to accurately 
localize and estimate the 3D bounding boxes of objects across varying 
room layouts, object scales, and occlusion levels, even in cluttered 
environments where multiple objects overlap in the image plane.

\noindent\textbf{Design Choice.} We adopt VLMs instead of 3D detectors as the Perceiver for two reasons. First, VLMs leverage open-world commonsense for amodal reasoning, while detectors degrade when objects are partially visible, which is a common case in cluttered indoor scenes. Second, 3D detectors output full 6-DoF poses, introducing per-instance orientation noise that disrupts our gravity-based canonicalization; VLMs produce approximately axis-aligned boxes that provide a cleaner input. Additionally, the choice of VLM extends naturally to language-guided 3D grounding and 3D spatial reasoning applications without architectural change.

\subsection{Action-based Refinement}
We visualize more resulting refined 3D layout generated by the LaP Planner 
across multiple scenes in Fig.~\ref{fig:supp_lapplanner}. We present the initialized 3D layout, layout after 1 iteration of refinement, and the final resulting structured 3D layout. Each example illustrates how the planner iteratively applies discrete actions to resolve physical deficiencies 
such as object interpenetration and floating objects, progressively 
improving the physical coherence of the layout over successive steps.

\section{Additional Quantitative Results}
\label{app:extra}

\noindent\textbf{Cross-dataset 3D grounding.} Tab.~\ref{tab:supp_cross_perceiver} reports the zero-shot 3D grounding results of our Perceiver 2B on 3D-FRONT and ARKitScenes, complementing Tab.~\ref{tab:crossdataset} in the main paper.

\begin{table}[ht!]
\centering
\small
\caption{Zero-shot 3D grounding on 3D-FRONT and ARKitScenes (500 images each).}
\label{tab:supp_cross_perceiver}
\begin{tabular}{l|l cccc}
\toprule
Dataset & Method & Reproj. IoU$\uparrow$ & Prec.{\tiny@(IoU=0.25)}$\uparrow$ & Prec.{\tiny@(IoU=0.5)}$\uparrow$ & Avg. DE$\downarrow$ \\
\midrule
\multirow{3}{*}{3D-FRONT}
 & N3D-VLM~\cite{wang2025n3d}        & 0.31 & 0.57 & 0.24 & 0.20 \\
 & DetAny3D~\cite{zhang2025detany3d} & 0.55 & 0.88 & \textbf{0.66} & 0.12 \\
 & Perceiver 2B (ours)               & \textbf{0.57} & \textbf{0.94} & \textbf{0.66} & \textbf{0.10} \\
\midrule
\multirow{3}{*}{ARKitScenes}
 & N3D-VLM~\cite{wang2025n3d}        & 0.23 & 0.40 & 0.16 & 0.31 \\
 & DetAny3D~\cite{zhang2025detany3d} & \textbf{0.44} & 0.79 & \textbf{0.54} & 0.29 \\
 & Perceiver 2B (ours)               & 0.42 & \textbf{0.81} & 0.49 & \textbf{0.22} \\
\bottomrule
\end{tabular}
\end{table}

\noindent\textbf{Composition of DPO rejected trajectories.} Our DPO rejected trajectories come from two sources (besides the \texttt{<STOP>}-only samples): Perceiver-initialized action sequences and synthetic perturbations of the ground truth. Tab.~\ref{tab:supp_dpo} disentangles their contributions on SUN RGB-D. Training only on Perceiver-initialized samples improves physical metrics but lowers Reproj. IoU, while synthetic-only training provides broader corrective patterns. Mixing both sources yields the best overall results, indicating that the two sources are complementary: synthetic data provides broad correction ability, while Perceiver-initialized data anchors refinement to realistic upstream errors. Note that evaluation always starts from Perceiver-initialized layouts.

\begin{table}[ht!]
\centering
\small
\caption{Ablation on the composition of the DPO rejected set (SUN RGB-D, Planner 2B).}
\label{tab:supp_dpo}
\begin{tabular}{l cccc}
\toprule
Rejected-set composition & Reproj. IoU$\uparrow$ & SVR (\%)$\downarrow$ & \# Collisions$\downarrow$ & Rot. Err.$\downarrow$ \\
\midrule
Before Ref.                          & 0.54 & 11.63 & 3.62 & 17.74 \\
Synthetic-perturbation only          & 0.59 & 5.68  & 2.43 & 11.26 \\
Perceiver-initialized only           & 0.51 & 9.83  & 3.39 & 11.90 \\
Mixed (20/20/60\% STOP/synth./Perc.) & 0.60 & \textbf{4.71} & 1.94 & 10.08 \\
Mixed (ours)                         & \textbf{0.65} & 4.85 & \textbf{1.46} & \textbf{9.05} \\
\bottomrule
\end{tabular}
\end{table}

\noindent\textbf{Robustness to detector errors.} Since our pipeline takes 2D boxes as input, we simulate common detector errors by perturbing the input boxes on SUN RGB-D: (a) \emph{jitter}, shifting each box edge by up to 20\% of the box width or height; (b) \emph{missing}, dropping each object with probability 20\% (counted as a missed prediction); and (c) \emph{mislabeling}, replacing each label with that of another object in the same image with probability 20\%. As shown in Tab.~\ref{tab:supp_robust}, performance degrades as expected but remains meaningful under all three perturbations.

\begin{table}[ht!]
\centering
\small
\caption{Robustness of the full pipeline (Perceiver 2B + Planner 2B) to perturbed input 2D boxes on SUN RGB-D.}
\label{tab:supp_robust}
\begin{tabular}{l cccc}
\toprule
Input 2D boxes & Reproj. IoU$\uparrow$ & Prec.{\tiny@(IoU=0.5)}$\uparrow$ & Avg. DE$\downarrow$ & SVR (\%)$\downarrow$ \\
\midrule
Without noise      & 0.65 & 0.63 & 0.07 & 4.85 \\
+ 20\% jitter      & 0.57 & 0.53 & 0.15 & 11.72 \\
+ 20\% missing     & 0.52 & 0.42 & 0.10 & 9.53 \\
+ 20\% mislabeling & 0.60 & 0.59 & 0.12 & 17.33 \\
\bottomrule
\end{tabular}
\end{table}

\section{Limitations and Future Work}
Our framework operates on a single image and uses off-the-shelf 
2D detectors as the entry point to 3D grounding; tighter coupling 
between detection and 3D reasoning, as well as extension to 
multi-view or video inputs, are natural next steps for resolving 
depth and occlusion ambiguities that are under-constrained from a 
single viewpoint. We focus on indoor scenes with well-defined 
support hierarchies, where the contact graph and gravity-aligned 
canonicalization are most informative; generalizing 
Layout-as-Policy to less structured environments is left for 
future work. Finally, our pipeline trains the Perceiver and 
Planner separately, and we believe joint optimization is a 
promising direction for further improving spatial reasoning, 
embodied planning, and interactive scene editing.

\section{Impact Statement}
\label{app:impact}
This paper focuses on the technical advancements in monocular 3D scene layout estimation using vision-language models. The proposed method aims to enable more accurate and robust spatial understanding from a single RGB image, with potential positive societal implications in domains such as robotics, embodied AI, augmented and virtual reality, interior design, and assistive technologies for visually impaired users. By improving the accessibility and reliability of 3D scene understanding from low-cost monocular sensors, the work may help broaden the reach of spatial AI applications to settings where specialized 3D sensing hardware is unavailable. As the paper primarily presents foundational technical research and does not involve the deployment of any system, we do not identify direct or immediate negative societal impacts specific to the proposed method itself.

%%%%%%%%%%%%%%%%%%%%%%%%%%%%%%%%%%%%%%%%%%%%%%%%%%%%%%%%%%%%

% \clearpage
% \input{checklist.tex}

\end{document}